\pdfoutput=1

\documentclass[11pt]{article}

\usepackage{EMNLP2023}

\usepackage{times}
\usepackage{latexsym}
\usepackage{booktabs}
\usepackage{inconsolata}
\usepackage{import}
\usepackage{microtype}
\usepackage{layout}
\usepackage{tabularx, makecell}
\usepackage{booktabs}
\usepackage{mathrsfs}
\usepackage{amssymb} 
\usepackage{url}
\usepackage{graphicx}
\usepackage{xspace,paralist}
\usepackage{times,latexsym}
\usepackage{amsmath}
\usepackage{appendix}
\usepackage{comment} 
\usepackage{enumitem}
\usepackage{makecell}
\usepackage{multirow}
\usepackage{xcolor}
\usepackage{graphicx}
\usepackage{cleveref}
\usepackage{tcolorbox}
\usepackage{adjustbox}
\usepackage{relsize}
\usepackage{todonotes}

\usepackage[T1]{fontenc}

\usepackage[utf8]{inputenc}

\usepackage{microtype}

\usepackage{inconsolata}
\usepackage{amsmath}
\usepackage{graphicx}
\usepackage{printlen}

\newcommand{\tabref}[2][]{Table#1~\ref{#2}\xspace}
\newcommand{\figref}[1]{Figure~\ref{#1}\xspace}
\newcommand{\secref}[1]{Section~\ref{#1}\xspace}
\newcommand{\appref}[1]{Appendix~\ref{#1}\xspace}

\newcommand{\model}[1]{\text{#1}\xspace}

\newcommand{\gptthreepointfive}{\model{GPT-3.5-Turbo}}

\newcommand{\gptfouro}{\model{GPT-4o}}

\newcommand{\llamatreb}{\model{LLaMA-3-8B}}

\newcommand{\empathicstories}{\model{EmpathicStories}}

\title{Can Machines Resonate with Humans?\\ Evaluating the Emotional and Empathic Comprehension of LMs}

\author{Muhammad Arslan Manzoor$^{1}$\thanks{\xspace\xspace Equal contribution.} \quad  
        Yuxia Wang$^{1}$$^*$  \quad  Minghan Wang$^2$  \quad Preslav Nakov$^{1}$ \\
  $^{1}$MBZUAI, Abu Dhabi, UAE \hspace{0.2em} $^{2}$University of Monash, Melbourne, Australia\\
  \texttt{\{muhammad.arslan, yuxia.wang, preslav.nakov\}@mbzuai.ac.ae}
  }
  
\begin{document}
\maketitle
\begin{abstract}


Empathy plays a pivotal role in fostering prosocial behavior, often triggered by the sharing of personal experiences through narratives. However, modeling empathy using NLP approaches remains challenging due to its deep interconnection with human interaction dynamics. Previous approaches, which involve fine-tuning language models (LMs) on human-annotated empathic datasets, have had limited success. In our pursuit of improving empathy understanding in LMs, we propose several strategies, including contrastive learning with masked LMs and supervised fine-tuning with large language models. While these methods show improvements over previous methods, the overall results remain unsatisfactory. To better understand this trend, we performed an analysis which reveals a low agreement among annotators. This lack of consensus hinders training and highlights the subjective nature of the task. We also explore the cultural impact on annotations. To study this, we meticulously collected story pairs in Urdu language and find that subjectivity in interpreting empathy among annotators appears to be independent of cultural background. Our systematic exploration of LMs' understanding of empathy reveals substantial opportunities for further investigation in both task formulation and modeling.

\end{abstract}

\section{Introduction}

\begin{figure}[!t]
    \centering
  \includegraphics[width=7.9cm,height=3.65cm]{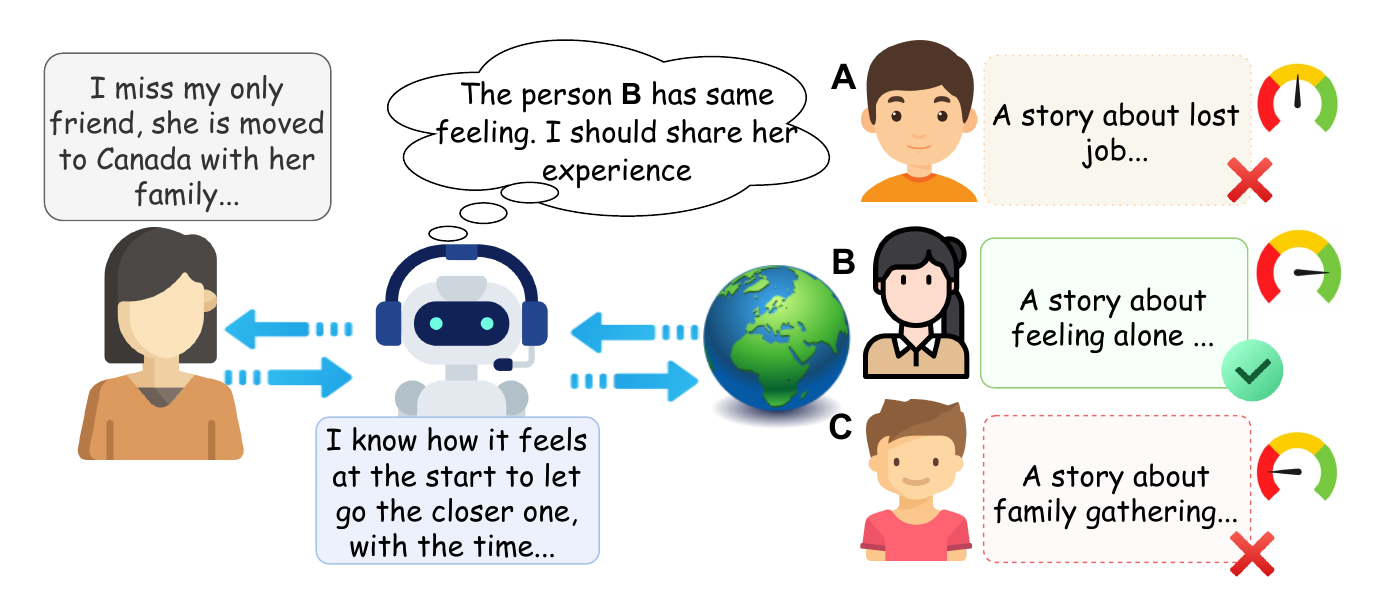}
  \caption{An \textit{ideal} interaction between users and a system. A chatbot can resonate with a human, and a search engine can retrieve stories of similar experience.}
  \label{fig:intro_fig}
  \vspace{-0.2em}
\end{figure}

With Large Language Models (LLMs) demonstrating impressive capabilities in generating naturally sounding answers over a broad range of human inquiries, more individuals turn to seek solutions and emotional comfort by interacting with LLM-supported chatbots~\cite{gpt4, chang2024survey}. They express thoughts, feelings and share their experiences, expecting deep understanding and sympathetic responses from a chatbot that can resonate with them, as shown in \figref{fig:intro_fig}~\citep{Berridge2023CompanionRT}. This requires LLMs to first fully understand the event, the emotions, and the empathy in the narratives, and then to respond appropriately~\cite{lin-etal-2023-ncuee}. 

\citet{shen-etal-2023-modeling} proposed the task of measuring empathic similarity which assesses the similarity between two narratives describing personal experiences across four aspects: event, emotion, moral, and empathy, using a numerical score ranging from 1 to 4 (see more in \secref{sec:task_dataset}). However, both fine-tuned LMs (BERT, RoBERTa) and few-shot prompted LLMs achieved low correlation with human annotations.
Error analysis on \citet{shen-etal-2023-modeling} shows that models can recognize dissimilar story pairs with scores in range of 1-2, but struggle to distinguish fine-grained differences between similar story pairs in range of 2.5-4 (see \secref{sec:casestudy}). This suggests that the nuanced patterns and relationships in similar pairs are not captured by current methods~\cite{wang-etal-2022-uncertainty}.

We hypothesize that the LMs used in these methods were primarily trained for semantic understanding tasks, rather than for emotion and empathy. They may capture the similarity between events, but they struggle to understand the complex social and emotional signals in human narratives \cite{Reimers2019SentenceBERTSE}, this is further discussed in following sections. Moreover, na\"{i}ve LLM prompts may not fully empower LLMs' reasoning ability to make correct judgements. 
To this end, we explore multiple strategies to improve the accuracy of empathic similarity, including contrastive learning of LMs, LLM reasoning and fine-tuning with and without Chain-of-Thought (CoT)~\citep{wei2022chain}.
We achieve around 5--10\% improvement in the Pearson and Spearman correlations. However, accuracy appears to be capped and cannot be improved beyond a certain value. 

This motivated us to speculate that gold labels might be problematic due to the subjective nature of judging emotion and empathy similarity. 
To investigate this, we randomly sampled ten story pairs from \empathicstories and asked eight annotators with different backgrounds to annotate the similarity in terms of event, emotion, moral and empathy (defined in \secref{sec:task_dataset}). 
Overall, low human agreement for all aspects was observed, especially moral, followed by empathy, emotion; the highest agreement was for the event. Interestingly, annotators from the same country/culture, particularly who are friends, had much higher agreement than others. We further collected a new Urdu dataset to explore the cultural impact on empathy similarity labeling, which revealed similar findings. 

In sum, this work presents three contributions: 
\begin{compactitem}
\item We explore various approaches to improve empathic similarity estimation, including the use of contrastive loss in LMs, reasoning and fine-tuning of LLMs.
\item We analyze the upper bound of correlation between model predictions and human gold labels. By gathering collective human opinions and measuring inter-annotator agreement, we reveal high disagreements in empathic labeling, highlighting the subjective nature of judging empathic similarity.
\item We collect a new Urdu dataset to investigate the impact of language and culture on empathic labeling. Our analysis shows that subjectivity in the labeling is independent of the cultural background of annotators.\footnote{Urdu dataset with individual similarity scores and modeling code are available at \url{https://github.com/yuxiaw/Empathic-Similarity}} 
\end{compactitem}

\section{Related Work}
\label{sec:related_work}

\paragraph{Emotion and Empathy in LMs:}
Many prior work focus on tasks of recognizing (mixed) emotions, scoring intensity, inferring and explaining person's emotional reactions~\citep{liu2024emollms, chen2024tombench}. 
In general domains, recent study illustrates that current LLMs are far better than human in generation, but fall short of understanding~\citep{West2023TheGA}. 
However, for empathy-demanding scenarios, understanding and resonating with support-seekers are even more important than outputting opinions (i.e., generation)~\cite{Buecker2021IsLI}.
In this work, we aim to examine language models capability in understanding and identifying nuanced difference of emotion and empathy, based on the task of \textit{empathic similarity}~\citep{shen-etal-2023-modeling}. 
We aim to improve the estimation accuracy by forcing models to discern and learn the underlying reasons for similarities between pairs of stories, via (1) enhancing LM-based sentence embeddings, and (2) empowering LLM reasoning capability. 

\paragraph{Sentence Embedding Enhancement:}
Many previous methods improve semantic embeddings by adjusting loss functions in training \cite{khan2022contrastive}.
SimCSE leverages contrastive loss~\cite{gao-etal-2021-simcse}, and ESimCSE applies momentum contrast strategy to increase the number of negative pairs involved in the loss calculation, showing notable improvements across multiple semantic textual similarity benchmarks~\citep{Wu2021ESimCSEES}. 
\citet{hubert2024treat} improved the knowledge graph embeddings by integrating semantic awareness into traditional loss functions like hinge loss. 
\citet{huang2024piccolo2} use a multi-task hybrid loss, incorporating both skeletal and semantic embedding into loss functions for micro-gesture classification.
Therefore, we investigate a variety of loss functions such as ContrastiveLoss, CosineSimilarityLoss, CoSENTLoss and AnglELoss, based on LMs including MPNet~\cite{song2020mpnet}, RoBERTa (base, large)~\cite{Liu2019RoBERTaAR}, DeBERTa (small, large)~\cite{he2020deberta} to improve sentence embedding in representing empathic features.



\paragraph{LLM Reasoning and Fine-tuning:}
Naive LLM prompts may fail to fully leverage LLMs' reasoning capabilities, leading to poor accuracy on \empathicstories dataset~\cite{shen-etal-2023-modeling}. Recent advancements in zero- and few-shot prompting techniques can boost performance such as Chain-of-Thought prompting (CoT) \cite{wei2022chain}, Least-to-Most prompting \cite{zhou2022least}, and search-based approaches like Tree-of-Thought (ToT) \cite{yao2024tree}. 
We prompt LLMs by CoT and also fine-tune LLMs with reasoning trajectories, encouraging LLMs to think over before making the final decision.


\section{Dataset and Error Analysis}
\label{sec:casestudy}
In this section, we first introduce the dataset used in this work and then we perform an error analysis for some baseline methods.

\subsection{Task and Dataset}
\label{sec:task_dataset}
In this work, we focus on the task of measuring the \textit{empathic similarity} between two narratives by a numerical score in the range of 1 to 4, with 1 representing totally dissimilar and 4 indicating extremely similar. 
Empathic similarity assesses how much the narrators of a pair of stories would empathize with one another, in which main event, emotion, and moral similarities of two stories are core features influencing the empathic similarity.

\textbf{Event} highlights the similarity of main events in two experiences, as people empathize more with experiences that are similar to their own.
\textbf{Emotional} reaction refers to how people emotionally respond to the experience. 
Individuals may have different feelings to the same experience, e.g., alone vs. happy for \textit{staying at home on weekend}.
\textbf{Moral} in this context emphasizes a higher level meaning abstracted by readers from a story, i.e., takeaway.

We used the \empathicstories dataset, which was created by \citet{shen-etal-2023-modeling}.
It consists of 1,500 unique stories and 2,000 story pairs, split into 1,500 pairs for training, 100 for development, and 400 for testing. Appendix \figref{englis_pair_example} shows the sample from EmpathicStories that consists of pair of stories and similarity scores on four labels.  \tabref{tab:error-analysis} shows the distribution of train, development, and test sets across different similarity ranges.
Each story has two versions, full and summary, in which the annotators assign labels based on the summary from four perspectives including  event, emotion, moral, and overall empathy. \figref{fig:rp_between_element_similarity} shows the correlation between the four similarity scores. We can see that the moral similarity has the highest correlation with empathy, followed by event and emotion.

\paragraph{Evaluation Measurements:} 
Pearson correlation ($r$) and Spearman rank correlation ($\rho$) are used to measure the performance of systems for predicting emotional similarity. This measures the linear correlation between the model outputs, the human annotations and the degree of monotonicity under ranking, respectively.
Mean square error (MSE) assesses the models' ability to get close to the gold standard. 
In a discrete setting, following \citet{shen-etal-2023-modeling}, both the predicted scores and the gold labels are binned into two classes thresholding by 2.5 --- label = 1 if score > 2.5; otherwise, it is 0. Then accuracy, precision, recall and macro-F1 are used.

\subsection{Error Analysis} 
\textbf{Baseline:} We reproduced the results of \citet{shen-etal-2023-modeling} by fine-tuning SBERT \cite{Reimers2019SentenceBERTSE} and BART-base~\cite{Lewis2019BARTDS} as shown in \tabref{tab:reproduce}, given that prompting \textit{davinci-text-003} and GPT-3.5-turbo had lower accuracy. We obtained a Spearman correlation $\rho$ similar to \citet{shen-etal-2023-modeling}, but with a lower F1.

\begin{table}[t!]
    \centering
    \resizebox{\columnwidth}{!}{%
    \begin{tabular}{@{}lccccccc@{}}
    \toprule
    \multicolumn{1}{l|}{\textbf{Splits$\rightarrow$}} & \textbf{Train} & \textbf{Dev.} & \multicolumn{1}{l|}{\textbf{Test.}} & \multicolumn{2}{c|}{\textbf{SBERT}} & \multicolumn{2}{c}{\textbf{BART}} \\
    \multicolumn{1}{l|}{\textbf{Label $\downarrow$}}  & ($1500$) & ($100$)  & \multicolumn{1}{l|}{($400$)} &
    \text{\#Dev} & \multicolumn{1}{c|}{\text{\#Test}} & \text{\#Dev} & \text{\#Test} \\ 
    \midrule
    \multicolumn{1}{l|}{\textbf{1}} & 85	& 2 & 15
    & \multicolumn{1}{|c}{0} & 0  & \multicolumn{1}{|c}{0} & 1 \\
    \multicolumn{1}{l|}{\textbf{1.5}} & 125	& 16 & 29 
    & \multicolumn{1}{|c}{0} & 0 & \multicolumn{1}{|c}{2} & 3 \\
    \multicolumn{1}{l|}{\textbf{2}} & 310	& 24 & 75 
    & \multicolumn{1}{|c}{1 } & 1  & \multicolumn{1}{|c}{11 } & 19 \\
    \multicolumn{1}{l|}{\textbf{2.5}} & 288	& 20 & 76 
    & \multicolumn{1}{|c}{4 } & 17  & \multicolumn{1}{|c}{15 } & 33  \\
    \multicolumn{1}{l|}{\textbf{3}} & 344& 21 & 103 
    & \multicolumn{1}{|c}{13 } & 49 & \multicolumn{1}{|c}{11 } & 55 \\
    \multicolumn{1}{l|}{\textbf{3.5}} & 262 	& 15 & 84 
    & \multicolumn{1}{|c}{12 } &70  & \multicolumn{1}{|c}{10 } & 57 \\
    \multicolumn{1}{l|}{\textbf{4}} & 86 	& 2 & 18
    & \multicolumn{1}{|c}{2 } & 18  & \multicolumn{1}{|c}{2 } & 15 \\
    \bottomrule
    \end{tabular}%
    }
    \caption{\empathicstories dataset distribution and the number of incorrect predictions on the development / test set by fine-tuned SBERT and BART baselines.}
    \label{tab:error-analysis}
    \vspace{-0.4em}
\end{table}

\paragraph{Results and Analysis:}
We regard a predicted score as a severely incorrect estimation when the absolute difference between the predicted score and the gold labels is greater than 1.0,\footnote{The cosine similarity scores based on SBERT and BART is in the range of [0,1], we scale them up by multiplying 4 to match the annotation range of 1-4.} the number of incorrect predictions over the development and the test sets is shown in \tabref{tab:error-analysis}. 

We found that the model excels at identifying dissimilar story pairs, but struggles with similar pairs with nuanced differences (not immediately obvious). Specifically, the model exhibits lower error rates in the score range of 1--2, and higher error rates in the range of 2.5--4. Particularly for 4, both models have a 100\% error rate on the test set.
We could attribute this to exposure bias during training. However, the number of training examples scoring between 2.5 and 4 is actually much larger than the number of dissimilar examples. 
This suggests that it is difficult for models to discern the subtle differences between similar story pairs, underlining a critical area for model improvement.


\paragraph{Hypothesis:}
We hypothesize that SBERT and BART are primarily pre-trained and fine-tuned to learn semantic features, in which emotional and empathetic features are under-represented.  
Despite the importance of understanding the semantics of the stories, this does not necessarily account for capturing the deep, empathetic connections between narratives, especially when these connections are subtler and more implicit than straightforward lexical or thematic commonalities. The failure to identify the connections between similar narratives in terms of empathy leads to poor performance on similar story pairs ranging from 2.5 to 4, tending to be recognized as dissimilar with a score between 1-2.
To alleviate these issues and to enhance the model performance on the empathetic similarity task, we propose a variety of strategies in \secref{sec:methods}.

\section{Enhancing Empathy Representation}
\label{sec:methods}
In order to improve the model's ability to recognize the connections between narratives that are semantically diverse yet empathically similar, we apply constrative losses in LM fine-tuning, and incorporate reasoning in LLM inference and fine-tuning.





\paragraph{Contrastive Learning of LMs:}
\label{sec:contrastive_learning}
We use a contrastive loss to enhance representation by bringing the embeddings of similar examples closer, while pushing such of dissimilar samples apart in the hidden space~\citep{gao-etal-2021-simcse}. This involves first grouping them into positive and negative pairs, and then applying a contrastive loss to learn patterns. In this task, we use annotated similarity scores to determine positive and negative pairs, setting a threshold of 2.5 as the boundary following the binning approach in \citet{shen-etal-2023-modeling}.

We explored various contrastive losses, including the ContrastiveLoss~\citep{DBLP:conf/cvpr/HadsellCL06}, AnglELoss~\citep{li2024angleoptimized} and CoSENTLoss\footnote{\href{https://sbert.net/docs/package/_reference/sentence/_transformer/losses.html}{www.sbert.net}}. We also re-implemented the approach using cosine similarity loss as a baseline.
We used masked language models (MLMs) including MPNet~\citep{song2020mpnet}, RoBERTa (base, large)~\citep{Liu2019RoBERTaAR} as text encoders leveraging their bidirectional encoding capabilities and computational efficiency. 


\paragraph{LLM Reasoning and Fine-tuning:}
\label{sec:llm_reasoning_method}
We explored the potential of LLMs for the task of empathic similarity in both inference and fine-tuning, with two prompting strategies: 
\begin{compactitem}
    \item \textbf{Score-Only}: With annotation guidelines, the LLM is prompted to predict only a score without any additional content.
    \item \textbf{Chain-of-Thought (CoT)}: The LLM is instructed to first provide an explanation and then to predict the scores. 
\end{compactitem}

For the CoT explanation, even though the human-annotated explanations were provided in \empathicstories, they were collected based on the story summary by \gptthreepointfive, which can be biased and inaccurate. Moreover, the distribution of human-written text is distinct from the distribution of LLM-generated reasons, which makes the reasons provided by the humans less useful for lowering the perplexity in LLM fine-tuning ~\citep{liu2024monotonic}.
To address this, we prompted an LLM to generate explanations guided by gold labels, leveraging LLM capability of causal reasoning. That is, to explain why two narratives have the similarity of \textit{[scores placeholder]} from the aspects of event, emotion, moral, and empathy. 

\paragraph{Gold Label-Guided Explanation:}
\label{sec:reasoning_synthesis}
Based on the pair of full stories with the ground truth similarity scores, we asked the LLM to analyze the story pair from dimensions such as theme, emotional content, characters, narrative structures, and overall empathy. The generated analysis served as the reasoning content in supervised fine-tuning (SFT).
We explored both full parameter and PEFT~\cite{hu2021lora} fine-tuning.
We used Llama3-70B-instruct to generate explanations by prompts in \figref{fig:explanationprompt} of \appref{app:explanation-generation}.

\section{Experiments}
\label{sec:exp}

We experimented with discriminative LMs and generative LLMs with zero-shot and fine-tuning.


\begin{table}[t!]
    \centering
    \resizebox{\columnwidth}{!}{
    \begin{tabular}{@{}lllccc@{}}
    \toprule
    \multicolumn{1}{l|}{\textbf{Train Label$\downarrow$}} & \multicolumn{1}{l|}{\textbf{Model}} & \multicolumn{1}{l|}{\textbf{Loss}} & \textbf{$r$} & \textbf{$\rho$} & \textbf{F1-macro} \\ \midrule
    \multicolumn{6}{c}{\texttt{\empathicstories Baselines} (Summary)} \\ \midrule
    \multicolumn{1}{l|}{\textbf{Empathy}} & \multicolumn{1}{l|}{SBERT} & \multicolumn{1}{l|}{MSE} & 0.359 & 0.352 & 0.713 \\
    \multicolumn{1}{l|}{\textbf{Empathy}} & \multicolumn{1}{l|}{BART} & \multicolumn{1}{l|}{MSE} & 0.342 & 0.344 & 0.701 \\
    \multicolumn{1}{l|}{\textbf{Empathy}} & \multicolumn{1}{l|}{GPT-3.5-turbo} & \multicolumn{1}{l|}{NA/5-shot} & 0.278 & 0.281 & 0.696 \\ \midrule
    \multicolumn{6}{c}{\texttt{Contrastive Loss of LMs} (Summary)} \\ \midrule
    \multicolumn{1}{l|}{\textbf{Empathy}} & \multicolumn{1}{l|}{RoBerta-base} & \multicolumn{1}{l|}{Cosine} & \textbf{0.404} & 0.388 & 0.608 \\
    \multicolumn{1}{l|}{\textbf{Empathy}} & \multicolumn{1}{l|}{Multi-qa-MPNet} & \multicolumn{1}{l|}{Cosine} & 0.400 & \textbf{0.395} & 0.647 \\ \midrule
    \multicolumn{1}{l|}{\textbf{Event}} & \multicolumn{1}{l|}{RoBerta-base} & \multicolumn{1}{l|}{Contrastive} & 0.318 & 0.309 & 0.634 \\
    \multicolumn{1}{l|}{\textbf{Event}} & \multicolumn{1}{l|}{Multi-qa-MPNet} & \multicolumn{1}{l|}{Contrastive} & 0.370 & 0.364 & 0.624 \\ \midrule
    \multicolumn{1}{l|}{\textbf{Emotion}} & \multicolumn{1}{l|}{RoBerta-base} & \multicolumn{1}{l|}{Cosine} & 0.377 & 0.371 & \textbf{0.650} \\
    \multicolumn{1}{l|}{\textbf{Emotion}} & \multicolumn{1}{l|}{RoBerta-large} & \multicolumn{1}{l|}{Cosine} & 0.393 & 0.388 & 0.611 \\ \midrule
    \multicolumn{1}{l|}{\textbf{Moral}} & \multicolumn{1}{l|}{RoBerta-base} & \multicolumn{1}{l|}{AnglELoss} & 0.326 & 0.323 & 0.649 \\
    \multicolumn{1}{l|}{\textbf{Moral}} & \multicolumn{1}{l|}{Multi-qa-MPNet} & \multicolumn{1}{l|}{Cosine} & 0.387 & 0.374 & 0.604 \\ \bottomrule
    \end{tabular}%
    }
\caption{Test set results for LM fine-tuned over annotations of \textbf{event}, \textbf{emotion}, \textbf{moral}, and overall \textbf{empathy} score and tested on \textbf{empathy} vs. baseline results obtained from \citet{shen-etal-2023-modeling}.}
    \label{tab:lm-finetuning}
\end{table}

\begin{table*}[ht!]
    \centering
    \resizebox{\textwidth}{!}{
    \begin{tabular}{@{}lcccccccccccccc@{}}
    \toprule
    \multicolumn{1}{l|}{\textbf{Testbed$\rightarrow$}} & \multicolumn{7}{c|}{\textbf{Summary}} & \multicolumn{7}{c}{\textbf{Full}} \\
    \multicolumn{1}{l|}{\textbf{Test Label$\downarrow$}} & \textbf{$r$} & \textbf{$\rho$} & \textbf{MSE}$\downarrow$ & \textbf{Acc} & \textbf{Prec} & \textbf{Recall} & \multicolumn{1}{c|}{\textbf{F1-macro}} & \textbf{$r$} & \textbf{$\rho$} & \textbf{MSE}$\downarrow$ & \textbf{Acc} & \textbf{Prec} & \textbf{Recall} & \textbf{F1-macro} \\ \midrule
    \multicolumn{15}{c}{\texttt{OpenAI-text-embedding-3-large}} \\ \midrule
    \multicolumn{1}{l|}{\textbf{Empathy}} & \multicolumn{1}{l}{0.336} & 0.329 & 1.510 & 0.505 & 0.633 & 0.517 & \multicolumn{1}{c|}{0.376} & \multicolumn{1}{l}{0.362} & 0.363 & 1.440 & 0.507 & 0.624 & 0.519 & 0.384 \\
    \multicolumn{1}{l|}{\textbf{Event}} & \multicolumn{1}{l}{\textbf{0.485}} & \textbf{0.465} & \textbf{0.620} & \textbf{0.780} & \textbf{0.738} & \textbf{0.542} & \multicolumn{1}{c|}{\textbf{0.522}} & \multicolumn{1}{l}{\textbf{0.488}} & \textbf{0.469} & \textbf{0.590} & \textbf{0.782} & \textbf{0.737} & \textbf{0.551} & \textbf{0.538} \\
    \multicolumn{1}{l|}{\textbf{Emotion}} & \multicolumn{1}{l}{0.392} & 0.388 & 1.310 & 0.550 & 0.582 & 0.510 & \multicolumn{1}{c|}{0.392} & \multicolumn{1}{l}{0.393} & 0.386 & 1.260 & 0.568 & 0.685 & 0.529 & 0.421 \\
    \multicolumn{1}{l|}{\textbf{Moral}} & \multicolumn{1}{l}{0.366} & 0.356 & 1.210 & 0.620 & \textbf{0.692} & \textbf{0.525} & \multicolumn{1}{c|}{\textbf{0.437}} & \multicolumn{1}{l}{0.395} & 0.403 & 1.140 & 0.618 & 0.651 & 0.524 & 0.440 \\ \midrule
    \multicolumn{15}{c}{\texttt{Llama-3-8B}} \\ \midrule
    \multicolumn{1}{l|}{\textbf{Empathy}} & \textbf{0.325} & 0.322 & \textbf{0.620} & \textbf{0.595} & 0.596 & 0.593 & \multicolumn{1}{c|}{\textbf{0.591}} & 0.324 & 0.308 & \textbf{0.520} & \textbf{0.590} & \textbf{0.595} & \textbf{0.592} & 0.588 \\
    \multicolumn{1}{l|}{\textbf{Event}} & \textbf{0.315} & \textbf{0.306} & 1.240 & 0.525 & 0.574 & 0.601 & \multicolumn{1}{c|}{0.509} & \textbf{0.342} & \textbf{0.312} & 0.900 & \textbf{0.660} & 0.617 & \textbf{0.659} & 0.611 \\
    \multicolumn{1}{l|}{\textbf{Emotion}} & 0.270 & 0.265 & 0.780 & 0.555 & 0.564 & 0.563 & \multicolumn{1}{c|}{0.554} & \textbf{0.317} & 0.294 & 0.600 & 0.595 & 0.590 & 0.588 & \textbf{0.588} \\
    \multicolumn{1}{l|}{\textbf{Moral}} & 0.319 & \textbf{0.323} & 0.830 & \textbf{0.600} & \textbf{0.622} & \textbf{0.623} & \multicolumn{1}{c|}{\textbf{0.600}} & 0.331 & \textbf{0.329} & 0.640 & 0.650 & \textbf{0.636} & 0.638 & \textbf{0.637} \\ \midrule
    \multicolumn{15}{c}{\texttt{Llama-3-70B}} \\ \midrule
    \multicolumn{1}{l|}{\textbf{Empathy}} & 0.405 & 0.403 & \textbf{0.620} & \textbf{0.635} & 0.661 & 0.630 & \multicolumn{1}{c|}{\textbf{0.614}} & 0.304 & 0.295 & \textbf{0.970} & \textbf{0.545} & 0.628 & 0.534 & \textbf{0.436} \\
    \multicolumn{1}{l|}{\textbf{Event}} & \textbf{0.427} & \textbf{0.431} & \textbf{1.280} & \textbf{0.480} & 0.623 & \textbf{0.639} & \multicolumn{1}{c|}{0.479} & \textbf{0.337} & \textbf{0.357} & \textbf{1.980} & \textbf{0.305} & \textbf{0.625} & \textbf{0.547} & \textbf{0.287} \\
    \multicolumn{1}{l|}{\textbf{Emotion}} & 0.387 & 0.374 & 0.770 & 0.565 & 0.605 & 0.585 & \multicolumn{1}{c|}{0.550} & 0.305 & 0.312 & 1.180 & 0.495 & 0.617 & 0.532 & 0.407 \\
    \multicolumn{1}{l|}{\textbf{Moral}} & 0.412 & 0.415 & 0.840 & 0.585 & \textbf{0.663} & \textbf{0.636} & \multicolumn{1}{c|}{\textbf{0.579}} & 0.305 & 0.320 & 1.320 & 0.455 & \textbf{0.659} & 0.545 & 0.391 \\ \midrule
    \multicolumn{15}{c}{\texttt{GPT-4o}} \\ \midrule
    \multicolumn{1}{l|}{\textbf{Empathy}} & 0.442 & 0.441 & 0.620 & 0.652 & 0.660 & 0.655 & \multicolumn{1}{c|}{0.650} & 0.350 & 0.373 & 0.650 & 0.640 & 0.640 & 0.640 & 0.640 \\
    \multicolumn{1}{l|}{\textbf{Event}} & \textbf{0.492} & \textbf{0.488} & \textbf{0.560} & \textbf{0.703} & 0.660 & \textbf{0.716} & \multicolumn{1}{c|}{0.659} & \textbf{0.414} & \textbf{0.424} & \textbf{0.710} & \textbf{0.605} & \textbf{0.615} & \textbf{0.660} & \textbf{0.579} \\
    \multicolumn{1}{l|}{\textbf{Emotion}} & 0.466 & 0.452 & 0.580 & 0.647 & 0.645 & 0.641 & \multicolumn{1}{c|}{0.641} & 0.360 & 0.371 & 0.660 & 0.620 & 0.622 & 0.622 & 0.620 \\
    \multicolumn{1}{l|}{\textbf{Moral}} & 0.476 & 0.481 & \textbf{0.560} & 0.698 & \textbf{0.685} & \textbf{0.687} & \multicolumn{1}{c|}{\textbf{0.686}} & 0.396 & \textbf{0.424} & \textbf{0.630} & \textbf{0.685} & \textbf{0.689} & \textbf{0.697} & \textbf{0.683} \\ \bottomrule
    \end{tabular}%
    }
    \caption{
    Zero-shot results of discriminative and generative models on the test set using summary and full story over four types of gold similarity scores: \textbf{empathy}, \textbf{event}, \textbf{emotion} and \textbf{moral}. Cosine similarity is scaled 1-4 by $\times$4 for discriminative models. Classification gold labels are binned by $\text{score} > 2.5$.}
    \label{tab:zs-compact-perf}
\end{table*}


\subsection{Discriminative LMs}
\label{sec:small_lm_ft}
We refer to discriminative models as smaller pre-trained LMs like RoBERTa, and sentence embedding models such as SBERT. In both zero-shot and fine-tuning settings, we calculated cosine similarity using the embeddings of two stories generated by either pre-trained or fine-tuned sentence encoders.



Each story pair in \empathicstories has four similarity scores: event, emotion, moral, and overall empathy. \citet{shen-etal-2023-modeling} only focused on fine-tuning and evaluating over empathy scores, under-investigating the other three aspects and their impact on the empathy similarity estimations.
In contrast, we experimented across all four labels, providing a comprehensive understanding of the model performance.

\subsubsection{Zero-Shot Evaluation}
We performed these experiments on both the full stories and on the summaries.
We calculated the cosine similarity based on the embeddings of a range of pre-trained text encoders including open-sourced \texttt{MiniLM}~\citep{wang2020minilm} and \texttt{MPNet}~\citep{song2020mpnet}, as well as close-sourced \texttt{OpenAI-text-embedding-3-large}~\citep{neelakantan2022text}, and then we evaluated their predictions against the gold labels from four perspectives.
\tabref{tab:lm-unsupervise-cosine} shows that (\emph{i})~the close-sourced model outperformed open-sourced models on both full and summary across all aspects in correlations, and (\emph{ii})~the cosine similarity scores for all models had the highest correlation with the event labels, indicating that semantics was the dominant feature captured in story representations. They were significantly notable than emotional and moral signals.



\subsubsection{Fine-Tuning with Contrastive Loss}
We examined the effectiveness of contrastive losses presented in \secref{sec:contrastive_learning} based on three LMs.
We fine-tuned them on the four types of gold labels separately and evaluated them using the overall empathy score as ground-truth based on pairs of story summary.\footnote{The full story always exceeds the maximum input length of these three LMs, resulting in poor performance.}
This enables us to assess the impact of each attribute on modeling empathic similarity. 

\tabref{tab:lm-finetuning} showcases the best two results over multiple LM$\times$Loss settings given the training label. We found that training on empathy similarity yields the highest correlation, followed by emotion, moral, and event across all settings. This suggests that empathy is more closely related to emotion and moral than to events. 
In terms of loss functions, CosineSimilarity loss consistently outperformed the rest, except for event as training labels, where Contrastive loss was the best. This highlights the robustness of CosineSimilarity loss across different scores. 
Among the three LMs, \texttt{Roberta-base} was always in the top-2 list, demonstrating robustness over all settings. \texttt{Roberta-large} and \texttt{Multi-qa-MPNet} outperformed in correlations when training with event, emotion, and moral.

These results suggest that, while some models and loss functions excel in specific metrics or training labels, \texttt{Roberta-base} with CosineSimilarity loss appears to be the most robust and effective combination for capturing the nuances of empathy. See the full results in \tabref{tab:lm-finetuning-ablation}. 

\subsection{Generative LLMs}
Unlike discriminative models to predict or measure a similarity score, LLMs explicitly generate scores in natural language.
We first evaluated LLMs in zero-shot setting to assess the inherent ability of LLMs to comprehend emotion and empathy, and then we fine-tuned the same LLMs to learn and to reason with the empathy in the stories. 

\subsubsection{Zero-Shot Evaluation}

We optimized the prompt used in \citet{shen-etal-2023-modeling} by specifying the differences between the four concepts, and by highlighting this in the system prompt instead of the user input, guiding the model to predict empathic similarity scores. 
Observing that the order of the stories affected the predicted scores, we generated two scores for each pair of stories by swapping the positions of stories A and B and then averaging the two scores as the final prediction.


\tabref{tab:zs-compact-perf} shows two major findings: 
(\emph{i})~Zero-shot LLMs have clear advantages over cosine similarity based on sentence embeddings. $r$=0.442 for \gptfouro vs. $r$=0.336 for \texttt{OpenAI-text-embedding} on story summary. The larger the model, the higher the correlations with the gold empathy scores.
This suggests that LLMs benefit from large-scale pre-training that empowers the distinguishability of nuanced empathic differences in narratives.
(\emph{ii}) Over all models and over the four types of scores, using the summaries is superior to using the full stories, which aligns with the fact that the gold labels are annotated based on the summary. 

\begin{table*}[ht!]
    \centering
    \resizebox{\textwidth}{!}{
    \begin{tabular}{l|ccccccc|ccccccc}
    \toprule
    \textbf{Testbed$\rightarrow$} & \multicolumn{7}{c|}{\textbf{Development Set}} & \multicolumn{7}{c}{\textbf{Test Set}} \\
    \textbf{SFT Strategy} & \textbf{$r$} & \textbf{$\rho$} & \textbf{MSE} & \textbf{Acc} & \textbf{Prec} & \textbf{Recall} & \textbf{F1-macro} & \textbf{$r$} & \textbf{$\rho$} & \textbf{MSE} & \textbf{Acc} & \textbf{Prec} & \textbf{Recall} & \textbf{F1-macro} \\ \midrule
    \textbf{LoRA-SFT-Score} & \textbf{0.470} & \textbf{0.476} & \textbf{0.497} & \textbf{0.790} & \textbf{0.759} & \textbf{0.786} & \textbf{0.768} & \textbf{0.307} & \textbf{0.320} & \textbf{0.643} & 0.600 & 0.602 & 0.603 & 0.599 \\
    \textbf{LoRA-SFT-COT} & 0.342 & 0.325 & 0.618 & 0.690 & 0.698 & 0.709 & 0.687 & 0.289 & 0.299 & 0.651 & \textbf{0.613} & \textbf{0.619} & \textbf{0.609} & \textbf{0.603} \\
    \textbf{Full-SFT-Score} & 0.209 & 0.189 & 0.627 & 0.630 & 0.640 & 0.632 & 0.625 & 0.028 & 0.039 & 0.773 & 0.507 & 0.506 & 0.506 & 0.504 \\ \midrule
    \end{tabular}%
    }
    \caption{\llamatreb SFT by three paradigms. LoRA-SFT-Score: score-only strategy tuned with LoRA. LoRA-SFT-COT: CoT strategy with LoRA. Full-SFT-Score tunes the model with full parameters on empathy scores.} 
    \label{tab:llama3-sft-results}
    \vspace{-0.4em}
\end{table*}

\subsubsection{Is SFT Helpful?}
To further investigate the potential of LLMs in this task, we performed supervised fine-tuning (SFT) with \llamatreb using the story summaries and the corresponding empathic similarity scores from the training set. In our experiments, we explored two different prompting strategies as introduced in \secref{sec:llm_reasoning_method}. For the \textbf{Score-Only} strategy, we implemented both full-parameter and LoRA~\citep{hu2021lora} SFT. For the \textbf{CoT} strategy, we only performed LoRA SFT due to the memory pressure of the longer sequence length.
For all strategies, we tuned the model for two epochs to avoid overfitting. 

\tabref{tab:llama3-sft-results}, shows that, compared to the zero-shot setting, SFT does not enhance the performance of \llamatreb regardless of the strategy. In some cases, fine-tuning even worsens the results. Moreover, the full-parameter SFT setting yields the poorest results, which is counter-intuitive. 
This motivated us to analyze why LLMs could not improve the performance of empathic similarity to $r$>0.5 at least, and why SFT even worsened the results.

\begin{table}[t]
\centering
\resizebox{0.88\columnwidth}{!}{%
\begin{tabular}{@{}l|cccc@{}}
\toprule
Gold$\rightarrow$ & y=1 & y=2 & y=3 & y=4 \\ 
\midrule
$P(Y|X_{y_{gold}=1})$ & 0.142 & 0.392 & 0.409 & 0.058 \\
$P(Y|X_{y_{gold}=2})$ & 0.140 & 0.386 & 0.418 & 0.056 \\
$P(Y|X_{y_{gold}=3})$ & 0.129 & 0.381 & 0.434 & 0.055 \\
$P(Y|X_{y_{gold}=4})$ & 0.132 & 0.378 & 0.428 & 0.062 \\ 
\midrule
$P(Y)$ & 0.140 & 0.399 & 0.404 & 0.057 \\ 
\bottomrule
\end{tabular}%
}
\caption{Probability estimated by fine-tuned \llamatreb over four groups binned by gold label y of story pairs. $X_{y_{gold}=1}$ is the group of examples whose gold empathy similarity is 1. $P(Y)$ represents empirical distribution of training labels. All values are computed on the training set.}
\label{tab:distributions}
\end{table}

\subsection{Understanding the Bottleneck}
\label{sec:understandbottleneck}
To understand what LLMs learned during fine-tuning and why they struggle with this task, we analyzed the predictions made by the fully fine-tuned model. We selected this model because it is specifically adapted to the task of empathic similarity estimation, minimizing interference from its pre-trained knowledge. 

Ideally, if the model learned how to predict the empathic similarity based on the input story pairs, its probability over four similarity classes [1,2,3,4] should be close to [1, 0, 0, 0] when the gold label of the input pair is 1, and to [0, 1, 0, 0] when the gold label of the input pair is 2, and so forth.
To observe the real predictive probability over each similarity class, we first grouped the training story pairs $X$ into four sets by their gold empathy labels $y$ (continuous scores are rounded to the nearest integer), denoted as $X_{y_{gold}=i}$ with $i \in [1,2,3,4]$.
Next, we computed the probability for each pair by applying the \textit{softmax} function to the logits of $\{1,2,3,4\}$ of the first predicted token, and then we averaged the probabilities across all pairs within each class, as shown in \tabref{tab:distributions}.

Regardless of the similarity class, the predicted probability kept the same as the empirical distribution of training data over four classes, i.e., $P(Y)$ = (0.140, 0.399, 0.404, 0.057), which is calculated by counting the percentage of story pairs falling into each class. 
That is, after fine-tuning, the model learned nothing about how to map the input $x \rightarrow y$, but merely the distribution of $P(Y)$. 
During inference, the model could not estimate the corresponding score conditioned on the input story pair, but sampled a similarity class based on the distribution of $P(Y)$ whatever the input pair was, leading to randomness on the development and the test sets (see confusion matrix of training set in \figref{fig:cm}).  

Based on these observations and previous findings in semantic textual similarity that models would be confused when training with ambiguous and subjective labels~\citep{wang-etal-2022-capture, wang-etal-2023-collective, wang-etal-2024-rethinking}, we speculate that the annotated empathic similarity scores are substantially subjective, i.e., that labels of high human disagreements hinder the model to learn distinguishable patterns.
This could also explain why LMs and LLMs could not improve the performance by a large margin despite our extensive efforts. 
The struggle of the models to overcome the bottleneck can be attributed to the complexity and the variability in the human interpretations of these abstract concepts. This underscores the need for a critical analysis of the dataset's quality and the inherent subjectivity of its labels. 
\section{Collective Human Opinions}
\label{sec:datasetanalysis}
To analyze the subjectivity in empathic similarity labeling, we first collected eight annotations for each pair of English stories. We then explored the impact of language and culture on empathy labeling by collecting and annotating Urdu story pairs.

\begin{table*}[t!]
    \centering
    \resizebox{\textwidth}{!}{
    \begin{tabular}{@{}lcccccccccccccccccccc@{}}
    \toprule
    \multicolumn{1}{l|}{\textbf{IAA Metric$\downarrow$}} & \textbf{S\_eve} & \textbf{S\_emo} & \textbf{S\_mor} & \textbf{S\_emp} & \textbf{F\_eve} & \textbf{F\_emo} & \textbf{F\_mor} & \textbf{F\_emp} & \textbf{Comb\_eve} & \multicolumn{1}{c|}{\textbf{Comb\_emo}} & \textbf{S\_eve} & \textbf{S\_emo} & \textbf{S\_mor} & \textbf{S\_emp} & \textbf{F\_eve} & \textbf{F\_emo} & \textbf{F\_mor} & \textbf{F\_emp} & \textbf{Comb\_eve} & \textbf{Comb\_emo} \\ 
    \midrule
    \multicolumn{10}{c}{\texttt{All annotators - 8}} & \multicolumn{10}{c}{English / Chinese Speakers (friends- 2)}\\ 
    \midrule
    \multicolumn{1}{l|}{\textbf{Pearson}} & 0.421 & 0.078 & 0.189 & 0.191 & 0.442 & 0.098 & 0.110 & -0.015 & 0.176  & \multicolumn{1}{c|}{0.141} & 0.735 & 0.322 & 0.192 & 0.148 & \textbf{0.848} & 0.311 & -0.086 & -0.241 & -0.302 & -0.153 \\
    \multicolumn{1}{l|}{\textbf{Spearman}} & 0.395 & 0.063 & 0.187 & 0.190 & 0.438 & 0.088 & 0.138 & 0.004 & 0.158 &  \multicolumn{1}{c|}{0.129} & 0.701 & 0.252 & 0.143 & 0.090 & \textbf{0.868} & 0.313 & -0.109 & -0.108 & -0.437 & -0.183 \\
    \multicolumn{1}{l|}{\textbf{KA}}  & 0.349 & 0.059 & 0.168 & 0.114 & 0.398 & 0.092 & 0.105 & 0.027 & 0.095 &   \multicolumn{1}{c|}{0.107} & 0.401 & 0.086 & 0.174 & -0.152 & 0.761 & 0.233 & -0.065 & -0.349 & -0.295 & -0.128 \\
    \multicolumn{1}{l|}{\textbf{Cohen Kappa}} & 0.124 & -0.015 & 0.067 & 0.047 & 0.157 & -0.024 & 0.030 & 0.026 & 0.059  & \multicolumn{1}{c|}{0.049} & 0.143 & -0.045 & 0.024 & -0.013 & 0.359 & -0.053 & 0.067 & -0.039 & -0.364 & -0.061 \\
    \midrule

    \multicolumn{10}{c}{\texttt{English / Urdu Speakers (colleagues - 4)}} & \multicolumn{10}{c}{ English / Urdu speakers (sisters - 2)}\\ 
    \midrule
    \multicolumn{1}{l|}{\textbf{Pearson}} & 0.266 & 0.037 & 0.089 & 0.093 & 0.322 & -0.017 & 0.234 & -0.039 & 0.137 & \multicolumn{1}{c|}{0.071} & \textbf{0.836} & 0.728 & 0.698 & 0.771 & 0.683 & 0.697 & 0.278 & 0.504 & 0.625 & 0.238 \\
    \multicolumn{1}{l|}{\textbf{Spearman}} & 0.235 & 0.028 & 0.101 & 0.078 & 0.295 & -0.013 & 0.194 & -0.075 & 0.133 & \multicolumn{1}{c|}{0.065} & \textbf{0.754} & 0.686 & 0.695 & 0.787 & 0.523 & 0.577 & 0.301 & 0.455 & 0.577 & 0.221 \\
    \multicolumn{1}{l|}{\textbf{KA}} & 0.235 & 0.051 & 0.050 & 0.026 & 0.311 & -0.008 & 0.187 & 0.057 & 0.098 & \multicolumn{1}{c|}{0.030}  & 0.724 & 0.476 & 0.709 & 0.773 & 0.659 & 0.560 & 0.273 & 0.486 & 0.317 & 0.195 \\
    \multicolumn{1}{l|}{\textbf{Cohen Kappa}} & 0.113 & 0.053 & -0.030 & 0.007 & 0.114 & -0.040 & -0.006 & -0.070 & 0.015 &  \multicolumn{1}{c|}{-0.016} & \textbf{0.429} & 0.167 & 0.375 & 0.412 & \textbf{0.412} & -0.045 & 0.143 & 0.403 & 0.138 & -0.014 \\
    \bottomrule
    \end{tabular}%
    }
    \caption{Agreement scores for four groups. The summary and the full story are abbreviated as S and F, Comb = S+F. Event, emotion, moral, and empathy are shortened as eve, emo, mor, and emp. KA refers to Krippendorff's Alpha.}
    \label{tab:agreement-score}
\end{table*}

\subsection{English Story Pair Annotation}
We sampled 10 story pairs from the development sets of \empathicstories, and we invited eight annotators to assign labels in three settings. 

\paragraph{Annotation Setup:}
\citet{shen-etal-2023-modeling} employed Amazon MTurk workers to annotate similarity scores based on the story summaries considering the heavy cognition load of understanding the full story. 
Upon performing qualitative analysis of both full stories and their summaries, we observed that summaries generated by GPT-3.5-turbo would dismiss many narratives presenting emotional changes (\textit{from depressed, to sad, finally turn to happy}), inner monologue (\textit{I feel alone after Mary left}) and details about other roles, only keeping the major events and the tone of the full stories. 
Yet, these details are very important to identify subtle feelings and they can affect empathy.
Model predictions based on the full stories also significantly differ from such obtained from the summaries as shown in \secref{sec:exp}. 
%
To this end, we collected annotations under three settings: 
\begin{compactitem} 
    \item \textbf{Continuous score on the summary:} The annotators assigned similarity scores for event, emotion, moral, and empathy as values ranging from 1 to 4, based on the story summaries.
    \item \textbf{Continuous score on the full story:} Same as above, but based on the full stories.  
    \item \textbf{Discrete class-label on full story:} Considering that continuous scores provide larger labeling space, which may exacerbate subjectivity and inner-annotator disagreement, the annotators are asked to rate event and emotion similarity using class labels: very similar (V), moderately similar (M), and not similar (N). Here, we did not rate moral and empathy given that the concepts were too abstract to perceive and rate.
\end{compactitem}
Based on the annotation results, we aim to measure and to analyze the subjectivity of the human opinions in different annotation setups, i.e., how the representation of the stories and the annotation scale influence inter-annotator agreement (IAA).
Pearson and Spearman correlation, Krippendorff's Alpha, and Cohen Kappa (convert to discrete) are used to measure IAA. We first calculated the agreement between each pair of annotators and then we averaged them.

\paragraph{Annotator Background:}
We had eight annotators, aged between 20 and 35 years, from diverse cultural and ethnic backgrounds, including two native Chinese speakers and six native Urdu speakers, with a balanced gender distribution.
All annotators are proficient in English, some majored in psychology with bachelor's degree and some are PhD and postdoc students in NLP. 
Training was performed before the formal annotations, which instructed the annotators how to rate the similarity score and highlighted the differences between event, emotion, moral, and empathy (guidelines in \figref{fig:instruction}).

\paragraph{Results and Analysis:}
\tabref{tab:agreement-score} shows the inter-annotator agreements. We can see that the agreement for event is the highest, followed by emotion, empathy, and moral.
This suggests that it is easier for the annotators to reach agreement on more concrete aspects such as event and emotion compared to more abstract concepts of moral and empathy.

The closer the relationship between two annotators, the higher the agreement for more subtle aspects.
Two Chinese friends (one male and one female) exhibited the highest correlation on judging event similarity based on the full story, but lower correlation on all other settings than the two sisters from Pakistan.
They even achieved $r/\rho$ >0.7 for moral and empathy when the average was less than 0.2 based on story summary.
Four Urdu speaker colleagues showed extremely low agreement with each other, with correlations around 0.3 for event and $r/\rho$<0.1 for other aspects.

This indicates that individuals who have closer interpersonal relationships with each other reach better agreement in interpretations of the stories, particular for subtle and abstract concepts.
%
Moreover, varying the agreement scores across different groups suggests that individual background and culture significantly influence how stories are perceived and annotated.
This subjectivity poses challenges for models to learn and to replicate human judgements, especially for moral and empathy. 
The agreement of collective human opinions in similarity score annotations also presents an upper bound for a model.

Comparing the three settings, we can see lower agreement when labeling with three classes than with continuous scores.
Moreover, using full stories yields higher correlations compared to using summaries.
This guides the setting of future annotations for empathic similarity, providing annotators with full stories and annotating concrete aspects like event and emotion using continuous labels.

\subsection{Urdu Dataset Construction}
Given that none of the annotators are native English speakers nor have grown up in a Western culture, bias might be introduced in labeling English stories pairs. We thus collected Urdu stories and asked Urdu native speakers to annotate them.
This aims to eliminate potential biases, and most importantly to investigate the impact of languages and culture on the empathy similarity labeling.
Urdu is widely spoken in many South Asian countries, including Pakistan, India, Bangladesh, and Nepal. Roman Urdu/Hindi, which uses the English alphabet to write Urdu or Hindi, is commonly used in these regions for communication, discussions, and sharing feelings on media platforms. Consequently, we collected stories written in Roman Urdu to capture the authentic expression and nuances of this widely used form of communication.

\begin{table}[t!]
    \centering
    \scalebox{0.78}{
    \begin{tabular}{@{}lcccc@{}}
    \toprule
    \multicolumn{1}{l|}{\textbf{IAA Metric$\downarrow$}} & \textbf{event} & \textbf{emotion} & \multicolumn{1}{c|}{\textbf{empathy}} & \textbf{overall} \\ 
    \midrule
    \multicolumn{1}{c}{\texttt{}}&\multicolumn{3}{c}{\texttt{All annotators (4)}} & \multicolumn{1}{|r}{GPT-4o}\\ 
    \midrule
    \multicolumn{1}{l|}{\textbf{Pearson}} & 0.308 & 0.381 &  \multicolumn{1}{c|}{0.422} & 0.425   \\
    \multicolumn{1}{l|}{\textbf{Spearman}} & 0.307 & 0.362 & \multicolumn{1}{c|}{0.392 } & 0.403 \\
    \multicolumn{1}{l|}{\textbf{KA}} & 0.214 & 0.316 & \multicolumn{1}{c|}{0.392} & 0.392 \\
    \multicolumn{1}{l|}{\textbf{Cohen Kappa}} & 0.133  & 0.169 & \multicolumn{1}{c|}{0.190} & 0.075 \\
       \bottomrule
    \end{tabular}%
    }
    \caption{Agreement scores for Urdu annotators on the Roman Urdu dataset. On the right, the empathy scores of 4 annotators are averaged correspondingly and compared to the overall GPT score.}
    \label{tab:urdu_agreement-score}
\end{table}

\begin{table}[t!]
\centering
\resizebox{\columnwidth}{!}{%
\begin{tabular}{@{}lccccccc@{}}
\toprule
\multicolumn{1}{l|}{\textbf{Gold Label$\downarrow$}} & \textbf{$r$} & \textbf{$\rho$} & \textbf{MSE}$\downarrow$ & \textbf{Acc} & \textbf{Prec} & \textbf{Recall} & \textbf{F1} \\ \midrule
\multicolumn{8}{c}{\texttt{OpenAI-text-embedding-3-large}} \\ \midrule
\multicolumn{1}{l|}{\textbf{Empathy}} & 0.486 & 0.527 & 1.610 & 0.296 & 0.609 & 0.562 & 0.289 \\
\multicolumn{1}{l|}{\textbf{Event}} & 0.539 & 0.548 & 1.040 & 0.399 & 0.651 & 0.564 & 0.372 \\
\multicolumn{1}{l|}{\textbf{Emotion}} & 0.493 & 0.529 & 1.600 & 0.292 & 0.607 & 0.562 & 0.286 \\ \midrule
\multicolumn{8}{c}{\texttt{Llama-3-8B}} \\ \midrule
\multicolumn{1}{l|}{\textbf{Empathy}} & 0.410 & 0.375 & 0.640 & 0.759 & 0.772 & 0.629 & 0.637 \\
\multicolumn{1}{l|}{\textbf{Event}} & 0.445 & 0.371 & 0.450 & 0.832 & 0.733 & 0.651 & 0.675 \\
\multicolumn{1}{l|}{\textbf{Emotion}} & 0.428 & 0.412 & 0.480 & 0.808 & 0.680 & 0.615 & 0.632 \\ \midrule
\multicolumn{8}{c}{\texttt{GPT-4o}} \\ \midrule
\multicolumn{1}{l|}{\textbf{Empathy}} & 0.715 & 0.725 & 0.330 & 0.797 &\textbf{ 0.762} & 0.757 & 0.759 \\
\multicolumn{1}{l|}{\textbf{Event}} & 0.772 & 0.762 & 0.310 & \textbf{0.842} & 0.760 & \textbf{0.848} & \textbf{0.786} \\
\multicolumn{1}{l|}{\textbf{Emotion}} & \textbf{0.774} & \textbf{0.774} & \textbf{0.310} & 0.818 & 0.732 & 0.807 & 0.754 \\ \bottomrule
\end{tabular}%
}
\caption{\textbf{Performance on the Urdu dataset} over three types of gold similarity scores: event, emotion, and empathy using cosine similarity ($v_a$, $v_b$) with OpenAI embedding and zero-shot LLMs.} 
\label{tab:urdu-llm-cosine}
\end{table}

\paragraph{Synthetic Story Pairs Generation:}
After manually checking the quality of the stories generated by GPT-4o (e.g., whether they are emotion-rich stories matching local culture), we collected 300 story pairs by instructing GPT-4o to generate story pairs with varying similarities given a diverse range of domains and topics (see prompts in \figref{fig:urdustoryprompt}). 
Each pair consisted of theme, content of two stories, an overall similarity score \textit{s} in [1,4], and a brief explaination of the score.


\paragraph{Human Annotations:}
Four Urdu native speakers were trained to annotate similarity scores for event, emotion and empathy, ranging from 1 to 4 with increments of 0.5; we excluded moral due to high ambiguity.
The IAA between the four annotators is shown in \tabref{tab:urdu_agreement-score}. The last column \textit{overall} represents the agreement between the averaged empathy scores of the four raters with the overall similarity score \textit{s} provided by GPT-4o during story generation. Interestingly, empathy achieved the highest IAA, followed by emotion and then event, which differs with the findings in English data: event>emotion>empathy. 

Agreement scores can be significantly influenced by shared interpersonal traits among annotators. Cultural factors also play a role, with closer interpersonal relationships leading to higher agreement scores. As demonstrated in the results, when annotators are more closely related, resonate with each other and have the same perception about the world, their agreement score is higher which may enhance the LMs capacity to learn cultural specific empathy. This suggests that, enhancing the machines’ understanding of empathy, stories must be collected and annotated within targeted countries or regions where there is a deep understanding of local demographic norms.
The correlations between the averaged empathy score and \gptfouro similarity are closer to those between annotators, suggesting the human-like judgement of \gptfouro.


\paragraph{LLM Estimations:}
Given the superior zero-shot performance obtained by OpenAI-text-embedding-3-large, \llamatreb, and \gptfouro on English dataset, we applied them on the Urdu story pairs in \tabref{tab:urdu-llm-cosine}. 
\gptfouro outperforms the other two models by a sizable margin. All of them have the highest scores on event, and then emotion and empathy, which is similar to the results for English, while the correlations of all aspects are much higher than for English across the three models. 


\section{Conclusion and Future Work}
We proposed a variety of strategies to enhance model performance on empathic similarity, including the use of contrastive losses on LMs, LLM reasoning and fine-tuning. Our experiments demonstrated 5--10\% improvements compared to baselines. 
However, our analysis revealed the subjective nature of empathic similarity between narratives.
Collective human annotations on both English and Urdu story pairs illustrated the low human agreement in empathy labeling, highlighting the inherent challenges of this task and indicating an upper bound that a model can reach.
The annotation of Urdu stories further exposed the cultural impact on empathic labeling. 

We find that many factors impact the inner annotator agreement in empathic similarity labeling, including the native language of annotators, the similarity of their background, experience, the closeness between them, and training process.  
Considering these factors that affect the correlations between annotators in a unified framework would be an interesting direction to explore in future work, especially the studies across multiple languages.
Furthermore, we will explore task reformulations to reduce the variability and subjectivity, and we will try more robust approaches to effectively model empathy in narratives.



\section*{Limitation}
The overall task inherently carries subjectivity. The gold labels are not standardized and vary based on individual backgrounds, demographics, perspectives, experiences, and surroundings. Cultural differences also play a significant role. Emotions are complex and varied, and each emotion can be expressed in multiple ways and at different intensities. Thus, gold labels are subjective, and stories themselves contain many nuances that may not be fully empathized by the annotators. Each emotion has a broad spectrum of intensity, rather than a binary state like happy or sad. 

\section*{Ethical Statement}

We adhere to ethical standards in data collection, annotation, and analysis. All human annotators were well informed about the task and provided their consent. We ensured diverse representation among the annotators to account for various cultural and demographic perspectives, aiming to minimize biases in empathic and emotional labeling. The datasets used, including those in Urdu, were collected and processed with respect for cultural sensitivity. We acknowledge the subjective nature of empathy and emotion analysis and have taken steps to highlight and to address these challenges in our study. Our work is committed to advancing understanding while maintaining ethical integrity and respect for the individuals whose data and annotations were used.

\bibliography{main}
\bibliographystyle{acl_natbib}

\clearpage
\onecolumn
\appendix
\section*{Appendix}



\section{Error Analysis}
\begin{figure}[ht!]
    \centering
    \includegraphics[scale=0.35]{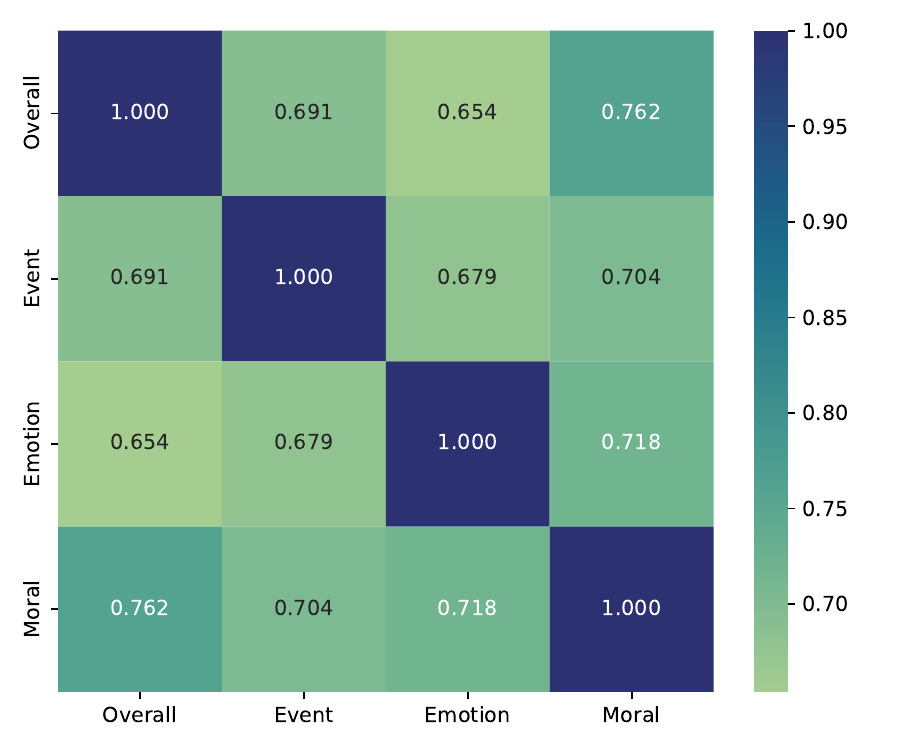}
    \includegraphics[scale=0.35]{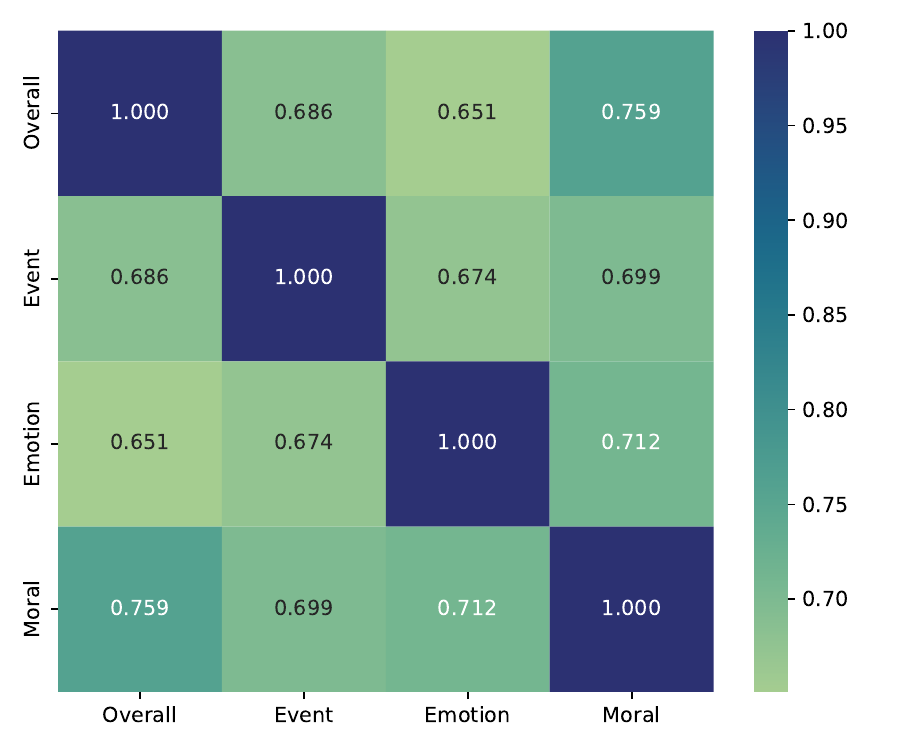}
    \caption{Pearson and Spearsman correlation between overall empathic similarity, event, emotion and moral similarity. Moral similarity has the highest correlation with the empathic, followed by event and emotion.}
	\label{fig:rp_between_element_similarity}
\end{figure}

\begin{figure}[ht!]
    \centering
    \includegraphics[width=0.65\textwidth, height=0.2\textheight, trim=26pt 10pt 0pt 0pt, clip]{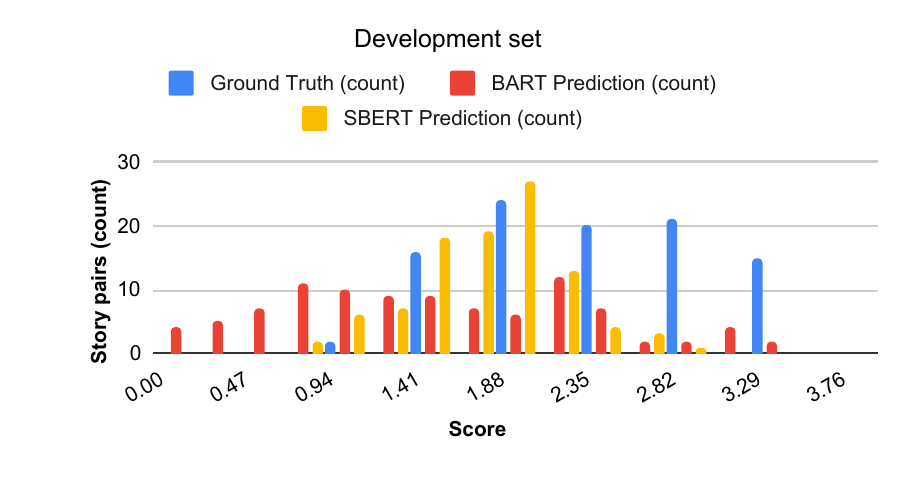} \\
    \includegraphics[width=0.65\textwidth, height=0.2\textheight, trim=12pt 15pt 0pt 0pt, clip]{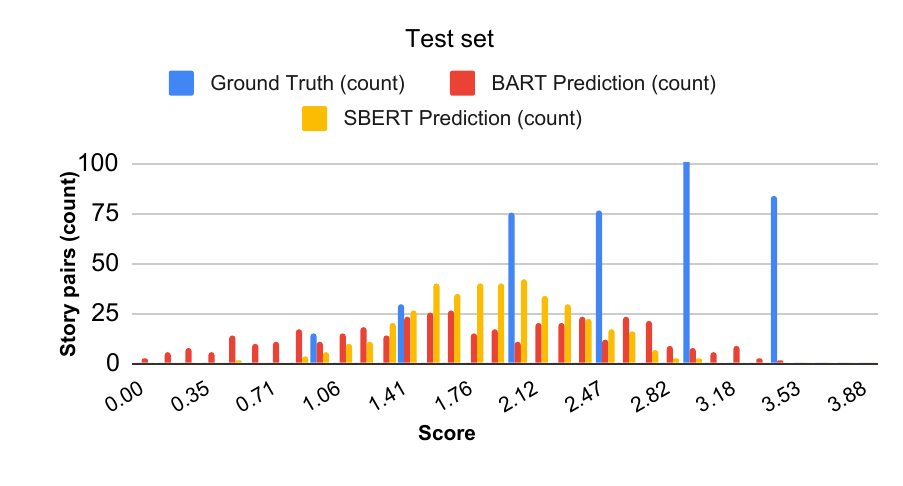}
    \caption{Dev/Test set empathic similarity distribution: predictions of BART vs. SBERT vs. ground truth.}
	\label{fig:dev_test_error}
\end{figure}

\figref{fig:rp_between_element_similarity} shows the correlation between the four similarity scores. Table \ref{tab:reproduce} exhibits the results authors mentioned in \cite{shen-etal-2023-modeling} and our reproduction results on EmpathicStories. Figure \ref{fig:dev_test_error} show the prediction count with SBERT and BART on test and development set.

\begin{table}[ht!]
\centering
\resizebox{0.45\columnwidth}{!}{%
\begin{tabular}{lccc}
\toprule
\textbf{Model} & \textbf{Macro-F1}$\uparrow$ & \textbf{MSE}$\downarrow$ & \textbf{$\rho \uparrow$} \\
\midrule
SBERT  (our rerun) & 0.560 & 0.224 & 0.317 \\
SBERT  (theirs) & 0.712 & -- & 0.352 \\
BART   (our rerun) & 0.579 & 0.240 & 0.389 \\
BART   (theirs) & 0.706 & -- & 0.344 \\
\bottomrule
\end{tabular}
}
\caption{Reproduced results vs. results presented in \citet{shen-etal-2023-modeling}.}
\label{tab:reproduce}
\end{table}

\clearpage
\section{Experiments}
\subsection{Gold-label Guided Explanation Generation}
\begin{figure}[ht!]
    \centering
        \includegraphics[scale=0.54]{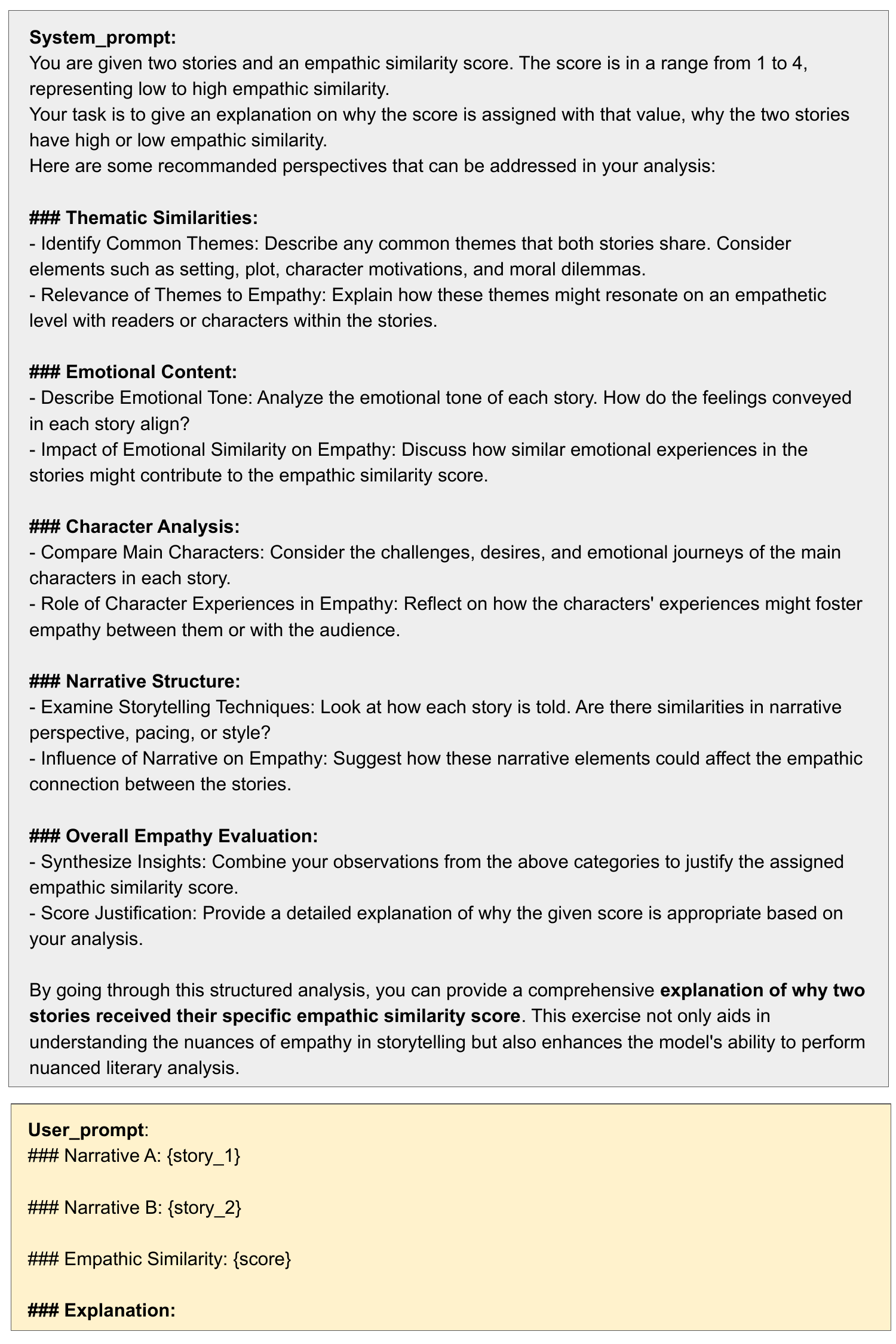}
    \caption{Gold-label Guided Explanation Generation Prompt using Llama3-70B-instruct.}
    \label{fig:explanationprompt}
\end{figure}

We used Llama3-70B-instruct with the prompts detailed in \figref{fig:explanationprompt} to analyze the story pairs across various dimensions, generating analysis that was subsequently used as reasoning content for supervised fine-tuning (SFT). 

\label{app:explanation-generation}

\subsection{Discriminative Models Results}

\begin{table*}[ht!]
    \centering
    \resizebox{\textwidth}{!}{
    \begin{tabular}{@{}lccccccclcccccc@{}}
    \toprule
    \multicolumn{1}{l|}{\textbf{Testbed$\rightarrow$}} & \multicolumn{7}{c|}{\textbf{Development Set}} & \multicolumn{7}{c}{\textbf{Test Set}} \\
    \multicolumn{1}{l|}{\textbf{Test Label$\downarrow$}} & \textbf{$r$} & \textbf{$\rho$} & \textbf{MSE}$\downarrow$ & \textbf{Acc} & \textbf{Prec} & \textbf{Recall} & \multicolumn{1}{c|}{\textbf{F1-macro}} & \multicolumn{1}{c}{\textbf{$r$}} & \textbf{$\rho$} & \textbf{MSE}$\downarrow$ & \textbf{Acc} & \textbf{Prec} & \textbf{Recall} & \textbf{F1-macro} \\ \midrule
    \multicolumn{15}{c}{\texttt{Multi-qa-MiniLM} (Summary)} \\ \midrule
    \multicolumn{1}{l|}{\textbf{Empathy}} & 0.204 & 0.201 & 0.0815 & 0.63 & 0.591 & 0.543 & \multicolumn{1}{c|}{0.513} & 0.220 & 0.213 & 0.095 & 0.550 & 0.590 & 0.557 & 0.508 \\
    \multicolumn{1}{l|}{\textbf{Event}} & \textbf{0.331} & \textbf{0.263} & \textbf{0.061} & \textbf{0.84} & \textbf{0.711} & \textbf{0.670} & \multicolumn{1}{c|}{\textbf{0.686}} & \textbf{0.311} & \textbf{0.294} & \textbf{0.056} & \textbf{0.740} & \textbf{0.623} & \textbf{0.609} & \textbf{0.614} \\
    \multicolumn{1}{l|}{\textbf{Emotion}} & 0.271 & 0.231 & 0.0792 & 0.72 & 0.643 & 0.578 & \multicolumn{1}{c|}{0.578} & 0.226 & 0.226 & 0.088 & 0.560 & 0.554 & 0.534 & 0.500 \\
    \multicolumn{1}{l|}{\textbf{Moral}} & 0.169 & 0.133 & 0.075 & 0.760 & 0.600 & 0.568 & \multicolumn{1}{c|}{0.574} & 0.216 & 0.214 & 0.084 & 0.610 & 0.575 & 0.550 & 0.533 \\ \midrule
    \multicolumn{15}{c}{\texttt{Multi-qa-MiniLM} (Full)} \\ \midrule
    \multicolumn{1}{l|}{\textbf{Empathy}} & 0.226 & 0.236 & 0.078 & 0.600 & 0.520 & 0.509 & \multicolumn{1}{c|}{0.466} & 0.151 & 0.13 & 0.092 & 0.520 & 0.543 & 0.527 & 0.475 \\
    \multicolumn{1}{l|}{\textbf{Event}} & \textbf{0.337} & \textbf{0.285} & \textbf{0.060} & \textbf{0.790} & \textbf{0.593} & \textbf{0.569} & \multicolumn{1}{c|}{\textbf{0.577}} & \textbf{0.268} & \textbf{0.239} & \textbf{0.058} & \textbf{0.715} & \textbf{0.584} & \textbf{0.574} & \textbf{0.578} \\
    \multicolumn{1}{l|}{\textbf{Emotion}} & 0.206 & 0.208 & 0.082 & 0.690 & 0.572 & 0.537 & \multicolumn{1}{c|}{0.524} & 0.226 & 0.226 & 0.088 & 0.560 & 0.554 & 0.534 & 0.500 \\
    \multicolumn{1}{l|}{\textbf{Moral}} & 0.169 & 0.139 & 0.075 & 0.760 & 0.600 & 0.568 & \multicolumn{1}{c|}{0.574} & 0.216 & 0.214 & 0.084 & 0.610 & 0.575 & 0.550 & 0.533 \\ \midrule
    \multicolumn{15}{c}{\texttt{Multi-qa-MPNet-base} (Summary)} \\ \midrule
    \multicolumn{1}{l|}{\textbf{Empathy}} & 0.170 & 0.140 & \textbf{0.066} & \textbf{0.440} & \textbf{0.554} & \textbf{0.528} & \multicolumn{1}{c|}{\textbf{0.405}} & 0.263 & 0.238 & \textbf{0.058} & \textbf{0.540} & 0.581 & 0.530 & 0.447 \\
    \multicolumn{1}{l|}{\textbf{Event}} & \textbf{0.298} & \textbf{0.280} & 0.133 & 0.270 & 0.515 & 0.513 & \multicolumn{1}{c|}{0.269} & \textbf{0.368} & \textbf{0.347} & 0.102 & 0.315 & 0.575 & 0.538 & 0.303 \\
    \multicolumn{1}{l|}{\textbf{Emotion}} & 0.224 & 0.204 & 0.086 & 0.370 & 0.544 & 0.525 & \multicolumn{1}{c|}{0.355} & 0.318 & 0.297 & 0.067 & 0.515 & \textbf{0.632} & \textbf{0.549} & \textbf{0.445} \\
    \multicolumn{1}{l|}{\textbf{Moral}} & 0.206 & 0.198 & 0.093 & 0.310 & 0.539 & 0.528 & \multicolumn{1}{c|}{0.306} & 0.274 & 0.256 & 0.075 & 0.445 & 0.571 & 0.527 & 0.392 \\ \midrule
    \multicolumn{15}{c}{\texttt{Multi-qa-MPNet-base} (Full)} \\ \midrule
    \multicolumn{1}{l|}{\textbf{Empathy}} & 0.145 & 0.127 & \textbf{0.076} & \textbf{0.440} & 0.6380 & 0.543 & \multicolumn{1}{c|}{\textbf{0.384}} & 0.299 & 0.288 & \textbf{0.065} & \textbf{0.530} & \textbf{0.645} & \textbf{0.518} & 0.389 \\
    \multicolumn{1}{l|}{\textbf{Event}} & \textbf{0.291} & \textbf{0.303} & 0.158 & 0.230 & 0.524 & 0.512 & \multicolumn{1}{c|}{0.223} & 0.349 & \textbf{0.346} & 0.132 & 0.250 & 0.500 & 0.500 & 0.218 \\
    \multicolumn{1}{l|}{\textbf{Emotion}} & 0.218 & 0.228 & 0.100 & 0.350 & 0.589 & 0.532 & \multicolumn{1}{c|}{0.320} & \textbf{0.350} & 0.333 & 0.078 & 0.465 & 0.537 & 0.504 & \textbf{0.347} \\
    \multicolumn{1}{l|}{\textbf{Moral}} & 0.231 & 0.227 & 0.108 & 0.290 & 0.614 & \textbf{0.550} & \multicolumn{1}{c|}{0.277} & 0.301 & 0.282 & 0.089 & 0.420 & 0.625 & 0.516 & 0.330 \\ \midrule
    \multicolumn{15}{c}{\texttt{OpenAI-text-embedding-3-large} (Summary)} \\ \midrule
    \multicolumn{1}{l|}{\textbf{Empathy}} & 0.335 & 0.315 & 1.280 & 0.630 & 0.813 & 0.513 & \multicolumn{1}{c|}{0.411} & 0.336 & 0.329 & 1.510 & 0.505 & 0.633 & 0.517 & 0.376 \\
    \multicolumn{1}{l|}{\textbf{Event}} & \textbf{0.437} & \textbf{0.411} & \textbf{0.600} & \textbf{0.820} & 0.414 & 0.494 & \multicolumn{1}{c|}{0.451} & \textbf{0.485} & \textbf{0.465} & \textbf{0.620} & \textbf{0.780} & \textbf{0.738} & \textbf{0.542} & \textbf{0.522} \\
    \multicolumn{1}{l|}{\textbf{Emotion}} & 0.394 & 0.350 & 1.130 & 0.720 & 0.859 & 0.517 & \multicolumn{1}{c|}{0.451} & 0.392 & 0.388 & 1.310 & 0.550 & 0.582 & 0.510 & 0.392 \\
    \multicolumn{1}{l|}{\textbf{Moral}} & 0.359 & 0.309 & 0.960 & 0.800 & \textbf{0.899} & \textbf{0.524} & \multicolumn{1}{c|}{\textbf{0.489}} & 0.366 & 0.356 & 1.210 & 0.620 & 0.692 & 0.525 & 0.437 \\ \midrule
    \multicolumn{15}{c}{\texttt{OpenAI-text-embedding-3-large} (Full)} \\ \midrule
    \multicolumn{1}{l|}{\textbf{Empathy}} & 0.303 & 0.309 & 1.400 & 0.620 & 0.561 & 0.505 & \multicolumn{1}{c|}{0.406} & 0.362 & 0.363 & 1.440 & 0.507 & 0.624 & 0.519 & 0.384 \\
    \multicolumn{1}{l|}{\textbf{Event}} & \textbf{0.385} & \textbf{0.409} & \textbf{0.680} & \textbf{0.810} & 0.413 & 0.488 & \multicolumn{1}{c|}{0.448} & \textbf{0.488} & \textbf{0.469} & \textbf{0.590} & \textbf{0.782} & \textbf{0.737} & \textbf{0.551} & \textbf{0.538} \\
    \multicolumn{1}{l|}{\textbf{Emotion}} & 0.345 & 0.361 & 1.260 & 0.710 & 0.607 & 0.510 & \multicolumn{1}{c|}{0.446} & 0.393 & 0.386 & 1.260 & 0.568 & 0.685 & 0.529 & 0.421 \\
    \multicolumn{1}{l|}{\textbf{Moral}} & 0.337 & 0.325 & 1.050 & 0.790 & \textbf{0.648} & \textbf{0.517} & \multicolumn{1}{c|}{\textbf{0.484}} & 0.395 & 0.403 & 1.140 & 0.618 & 0.651 & 0.524 & 0.440 \\ \bottomrule
    \end{tabular}%
    }
    \caption{\textbf{Cosine similarity($v_a$, $v_b$) across three sentence embedding models.} Results on dev and test sets over four type of gold similarity scores: \textbf{empathy}, \textbf{event}, \textbf{emotion} and \textbf{moral} based on full story and summary. Cosine similarity is normalized to the scale 1-4 by $\times$4. Classification gold labels are binned by $\text{score} > 2.5$.}
    \label{tab:lm-unsupervise-cosine}
\end{table*}

\begin{table*}[ht!]
    \centering
    \resizebox{\textwidth}{!}{
    \begin{tabular}{@{}cc|cccccc|cccccc@{}}
    \toprule
    \multicolumn{1}{l|}{\textbf{Model}} & \multicolumn{1}{l|}{\textbf{Loss}} & 
    \textbf{$r$} & \textbf{$\rho$}  & \textbf{Acc} & \textbf{Prec} & \textbf{Recall} & \multicolumn{1}{c|}{\textbf{F1-macro}} & \textbf{$r$} & \textbf{$\rho$}  & \textbf{Acc} & \textbf{Prec} & \textbf{Recall} & \textbf{F1-macro} \\ 
    \midrule
    \multicolumn{2}{c}{\texttt{}} & \multicolumn{6}{c}{\texttt{\textbf{Empathy}}} & \multicolumn{6}{c}{\texttt{\textbf{Event}}} \\ 
    \midrule
    \multicolumn{1}{l|}{\textbf{RoBerta-base}} & \multicolumn{1}{l|}{\textbf{AnglELoss}}  & 0.355 & 0.351 & 0.640 & 0.642 & 0.641 & 0.640 & 0.254 & 0.248 & 0.593 & 0.608 & 0.597 & 0.583 \\
    \multicolumn{1}{l|}{\textbf{RoBerta-base}} & \multicolumn{1}{l|}{\textbf{Cosine}} & \textbf{0.404} & 0.389 & 0.620 & 0.629 & 0.616 & 0.609 & 0.337 & 0.315 & 0.565 & 0.622 & 0.573 & 0.520 \\ 
    \multicolumn{1}{l|}{\textbf{RoBerta-base}} & \multicolumn{1}{l|}{\textbf{Contrastive}}  & 0.331 & 0.319 & 0.637 & 0.639 & 0.636 & 0.634 & 0.318 & 0.309 & \textbf{0.635} & 0.637 & \textbf{0.636} & \textbf{0.635}\\
    
   \multicolumn{1}{l|}{\textbf{RoBerta-large}} & \multicolumn{1}{l|}{\textbf{ConSENT}} &  0.291 & 0.298 & 0.603 & 0.611 & 0.605 & 0.599 &0.281 & 0.299 & 0.590 & 0.602 & 0.594 & 0.583\\
    \multicolumn{1}{l|}{\textbf{RoBerta-large}} & \multicolumn{1}{l|}{\textbf{Cosine}} & 0.373 & 0.346 & 0.608 & 0.612 & 0.604 & 0.599 & \textbf{0.353} & 0.350 & 0.593 & \textbf{0.653} & 0.600 & 0.557 \\
    
    \multicolumn{1}{l|}{\textbf{Multi-qa-MPNet}} & \multicolumn{1}{l|}{\textbf{Cosine}} & 0.400 & \textbf{0.396} & \textbf{0.647} & \textbf{0.648} & \textbf{0.648} & \textbf{0.647} & 0.343 & 0.346 & 0.573 & 0.643 & 0.581 & 0.525\\ 
    \multicolumn{1}{l|}{\textbf{Multi-qa-MPNet}} & \multicolumn{1}{l|}{\textbf{Contrastive}}  & 0.358 & 0.347 & 0.615 & 0.615 & 0.614 & 0.613 &	0.371 & \textbf{0.365} & 0.625 & 0.629 & 0.627 & 0.624 \\
    \midrule
    \multicolumn{2}{c}{\texttt{}} & \multicolumn{6}{c}{\texttt{\textbf{Emotion}}} & \multicolumn{6}{c}{\texttt{\textbf{Moral}}} \\
    \midrule
    \multicolumn{1}{l|}{\textbf{RoBerta-base}} & \multicolumn{1}{l|}{\textbf{AnglE}}  & 0.292 & 0.279 & 0.580 & 0.581 & 0.581 & 0.580  & 0.326 & 0.324 & \textbf{0.650} & 0.650 & 0.650 & \textbf{0.650} \\
    \multicolumn{1}{l|}{\textbf{RoBerta-base}} & \multicolumn{1}{l|}{\textbf{Cosine}} & 0.378 & 0.371 & \textbf{0.652} & \textbf{0.654} & \textbf{0.651} & \textbf{0.650} & 0.339 & 0.334 & 0.645 & 0.645 & 0.645 & 0.645 \\ 
   \multicolumn{1}{l|}{\textbf{RoBerta-base}} & \multicolumn{1}{l|}{\textbf{Contrastive}}  & 0.335 & 0.321 & 0.618 & 0.626 & 0.614 & 0.607 & 0.346 & 0.331 & 0.642 & 0.645 & 0.644 & 0.642\\
    \multicolumn{1}{l|}{\textbf{RoBerta-large}} & \multicolumn{1}{l|}{\textbf{CoSENT}}  & 0.303 & 0.287 & 0.585 & 0.585 & 0.585 & 0.585 & 0.287 & 0.311 & \textbf{0.650} & \textbf{0.651} & \textbf{0.651} & 0.650\\
    \multicolumn{1}{l|}{\textbf{RoBerta-large}} & \multicolumn{1}{l|}{\textbf{Cosine}}  & \textbf{0.394} & \textbf{0.388} & 0.620 & 0.626 & 0.617 & 0.611  & \textbf{0.387} & 0.373 & 0.640 & 0.640 & 0.640 & 0.640 \\
    \multicolumn{1}{l|}{\textbf{Multi-qa-MPNet}} & \multicolumn{1}{l|}{\textbf{Cosine}} & 0.367 & 0.359 & 0.613 & 0.612 & 0.611 & 0.611 & \textbf{0.387} & \textbf{0.374} & 0.608 & 0.615 & 0.610 & 0.605 \\ 
   \multicolumn{1}{l|}{\textbf{Multi-qa-MPNet}} & \multicolumn{1}{l|}{\textbf{Contrastive}}  & 0.339 & 0.325 & 0.610 & 0.610 & 0.609 & 0.608  & 0.368 & 0.362 & 0.608 & 0.616 & 0.610 & 0.604\\
    \bottomrule
    \end{tabular}%
    }
\caption{Performance of LMs fine-tuned based on annotations of \textbf{event}, \textbf{emotion} and \textbf{moral} \& overall \textbf{empathy} similarity scores respectively. Note that we consistently evaluate against empathy similarity score (test gold labels) though training with labels from four aspects.}
    \label{tab:lm-finetuning-ablation}
\end{table*}

\clearpage
\subsection{Generative LLMs Results}
\paragraph{Zero-shot Generative LLMs.} \tabref{tab:zs-compact-perf} demonstrates the results of zero-shot generative LLMs including Llama-3-8B, Llama-3-70B and GPT-4o on test set, \tabref{tab:llama3-zs-results} provides the additional results over development set.

\begin{table*}[ht!]
    \centering
    \resizebox{\textwidth}{!}{
    \begin{tabular}{@{}lcccccccccccccc@{}}
    \toprule
    \multicolumn{1}{l|}{\textbf{Testbed$\rightarrow$}} & \multicolumn{7}{c|}{\textbf{Development Set}} & \multicolumn{7}{c}{\textbf{Test Set}} \\
    \multicolumn{1}{l|}{\textbf{Test Label$\downarrow$}} & \textbf{$r$} & \textbf{$\rho$} & \textbf{MSE} & \textbf{Acc} & \textbf{Prec} & \textbf{Recall} & \multicolumn{1}{c|}{\textbf{F1-macro}} & \textbf{$r$} & \textbf{$\rho$} & \textbf{MSE} & \textbf{Acc} & \textbf{Prec} & \textbf{Recall} & \textbf{F1-macro} \\ \midrule
    \multicolumn{15}{c}{\texttt{Llama-3-8B} (Summary)} \\ 
    \midrule
    \multicolumn{1}{l|}{\textbf{Empathy}} & \textbf{0.494} & \textbf{0.500} & \textbf{0.550} & \textbf{0.740} & \textbf{0.751} & \textbf{0.765} & \multicolumn{1}{c|}{\textbf{0.738}} & \textbf{0.325} & 0.322 & \textbf{0.620} & 0.595 & 0.596 & 0.593 & 0.591 \\
    \multicolumn{1}{l|}{\textbf{Event}} & 0.468 & 0.465 & 1.340 & 0.570 & 0.597 & 0.671 & \multicolumn{1}{c|}{0.531} & 0.315 & 0.306 & 1.240 & 0.525 & 0.574 & 0.601 & 0.509 \\
    \multicolumn{1}{l|}{\textbf{Emotion}} & 0.459 & 0.466 & 0.780 & 0.630 & 0.648 & 0.678 & \multicolumn{1}{c|}{0.619} & 0.270 & 0.265 & 0.780 & 0.555 & 0.564 & 0.563 & 0.554 \\
    \multicolumn{1}{l|}{\textbf{Moral}} & 0.411 & 0.400 & 0.900 & 0.590 & 0.614 & 0.671 & \multicolumn{1}{c|}{0.563} & 0.319 & \textbf{0.323} & 0.830 & \textbf{0.600} & \textbf{0.622} & \textbf{0.623} & \textbf{0.600} \\ \midrule
    \multicolumn{15}{c}{\texttt{Llama-3-8B} (Full)} \\ \midrule
    \multicolumn{1}{l|}{\textbf{Empathy}} & \textbf{0.326} & \textbf{0.351} & \textbf{0.540} & 0.670 & \textbf{0.646} & \textbf{0.637} & \multicolumn{1}{c|}{\textbf{0.640}} & 0.324 & 0.308 & \textbf{0.520} & 0.590 & 0.595 & 0.592 & 0.588 \\
    \multicolumn{1}{l|}{\textbf{Event}} & 0.208 & 0.212 & 1.220 & \textbf{0.680} & 0.577 & 0.620 & \multicolumn{1}{c|}{0.573} & \textbf{0.342} & 0.312 & 0.900 & \textbf{0.660} & 0.617 & \textbf{0.659} & 0.611 \\
    \multicolumn{1}{l|}{\textbf{Emotion}} & 0.235 & 0.233 & 0.780 & 0.660 & 0.600 & 0.608 & \multicolumn{1}{c|}{0.603} & 0.317 & 0.294 & 0.600 & 0.595 & 0.590 & 0.588 & 0.588 \\
    \multicolumn{1}{l|}{\textbf{Moral}} & 0.165 & 0.147 & 0.860 & 0.600 & 0.502 & 0.502 & \multicolumn{1}{c|}{0.493} & 0.331 & \textbf{0.329} & 0.640 & 0.650 & \textbf{0.636} & 0.638 & \textbf{0.637} \\ 
    \midrule
    \multicolumn{15}{c}{\texttt{Llama-3-70B} (Summary)} \\ \midrule
    \multicolumn{1}{l|}{\textbf{Empathy}} & \textbf{0.450} & \textbf{0.459} & \textbf{0.610} & \textbf{0.700} & \textbf{0.705} & \textbf{0.717} & \multicolumn{1}{c|}{\textbf{0.697}} & 0.405 & 0.403 & \textbf{0.620} & \textbf{0.635} & 0.661 & 0.630 & \textbf{0.614} \\
    \multicolumn{1}{l|}{\textbf{Event}} & 0.420 & 0.403 & 1.400 & 0.570 & 0.583 & 0.647 & \multicolumn{1}{c|}{0.524} & \textbf{0.427} & \textbf{0.431} & 1.280 & 0.480 & 0.623 & \textbf{0.639} & 0.479 \\
    \multicolumn{1}{l|}{\textbf{Emotion}} & 0.416 & 0.435 & 0.840 & 0.630 & 0.639 & 0.668 & \multicolumn{1}{c|}{0.616} & 0.387 & 0.374 & 0.770 & 0.565 & 0.605 & 0.585 & 0.550 \\
    \multicolumn{1}{l|}{\textbf{Moral}} & 0.369 & 0.369 & 0.960 & 0.610 & 0.622 & 0.683 & \multicolumn{1}{c|}{0.579} & 0.412 & 0.415 & 0.840 & 0.585 & \textbf{0.663} & 0.636 & 0.579 \\ \midrule
    \multicolumn{15}{c}{\texttt{Llama-3-70B} (Full)} \\ \midrule
    \multicolumn{1}{l|}{\textbf{Empathy}} & 0.264 & 0.284 & \textbf{1.110} & \textbf{0.460} & \textbf{0.621} & \textbf{0.554} & \multicolumn{1}{c|}{\textbf{0.421}} & 0.304 & 0.295 & \textbf{0.970} & \textbf{0.545} & 0.628 & 0.534 & \textbf{0.436} \\
    \multicolumn{1}{l|}{\textbf{Event}} & \textbf{0.380} & \textbf{0.402} & 2.220 & 0.270 & 0.549 & 0.537 & \multicolumn{1}{c|}{0.268} & \textbf{0.337} & \textbf{0.357} & 1.980 & 0.305 & 0.625 & \textbf{0.547} & 0.287 \\
    \multicolumn{1}{l|}{\textbf{Emotion}} & 0.344 & 0.333 & 1.380 & 0.390 & 0.617 & 0.560 & \multicolumn{1}{c|}{0.372} & 0.305 & 0.312 & 1.180 & 0.495 & 0.617 & 0.532 & 0.407 \\
    \multicolumn{1}{l|}{\textbf{Moral}} & 0.288 & 0.262 & 1.580 & 0.330 & 0.619 & 0.576 & \multicolumn{1}{c|}{0.325} & 0.305 & 0.320 & 1.320 & 0.455 & \textbf{0.659} & 0.545 & 0.391 \\
    \midrule
    \multicolumn{15}{c}{\texttt{GPT-4o} (Summary)} \\ 
    \midrule
    \multicolumn{1}{l|}{\textbf{Empathy}} & 0.482 & 0.479 & 0.620 & 0.740 & \textbf{0.726} & 0.709 & \multicolumn{1}{c|}{\textbf{0.714}} & 0.442 & 0.441 & 0.620 & 0.652 & 0.660 & 0.655 & 0.650 \\
    \multicolumn{1}{l|}{\textbf{Event}} & 0.474 & 0.457 & 0.660 & 0.710 & 0.605 & 0.662 & 
    \multicolumn{1}{c|}{0.608} & \textbf{0.492} & \textbf{0.488} & \textbf{0.560} & \textbf{0.703} & 0.660 & \textbf{0.716} & 0.659 \\
    \multicolumn{1}{l|}{\textbf{Emotion}} & \textbf{0.518} & \textbf{0.514} & \textbf{0.590} & \textbf{0.750} & 0.700 & \textbf{0.712} & 
    \multicolumn{1}{c|}{0.705} & 0.466 & 0.452 & 0.580 & 0.647 & 0.645 & 0.641 & 0.641 \\
    \multicolumn{1}{l|}{\textbf{Moral}} & 0.423 & 0.416 & 0.650 & 0.710 & 0.621 & 0.659 & 
    \multicolumn{1}{c|}{0.628} & 0.476 & 0.481 & \textbf{0.560} & 0.698 & \textbf{0.685} & 0.687 & \textbf{0.686} \\
    \midrule
    \multicolumn{15}{c}{\texttt{GPT-4o} (Full)} \\ 
    \midrule
    \multicolumn{1}{l|}{\textbf{Empathy}} & 0.351 & 0.351 & \textbf{0.630} & \textbf{0.710} & \textbf{0.697} & \textbf{0.705} 
    & \multicolumn{1}{c|}{\textbf{0.699}} & 0.350 & 0.373 & 0.650 & 0.640 & 0.640 & 0.640 & 0.640 \\
    \multicolumn{1}{l|}{\textbf{Event}} & 0.385 & 0.367 & 0.850 & 0.660 & 0.616 & 0.702 
    & \multicolumn{1}{c|}{0.595} & \textbf{0.414} & \textbf{0.424} & 0.710 & 0.605 & 0.615 & 0.660 & 0.579 \\
    \multicolumn{1}{l|}{\textbf{Emotion}} & \textbf{0.397} & \textbf{0.385} & 0.670 & 0.680 & 0.654 & 0.683 
    & \multicolumn{1}{c|}{0.653} & 0.360 & 0.371 & 0.660 & 0.620 & 0.622 & 0.622 & 0.620 \\
    \multicolumn{1}{l|}{\textbf{Moral}} & 0.271 & 0.236 & 0.780 & 0.660 & 0.622 & 0.680 
    & \multicolumn{1}{c|}{0.609} & 0.396 & \textbf{0.424} & \textbf{0.630} & \textbf{0.685} & \textbf{0.689} & \textbf{0.697} & \textbf{0.683} \\
    \bottomrule
    \end{tabular}%
    }
    \caption{Zero-shot Generative LLMs results using the full story vs. story summary, over four types of gold similarity scores: empathy, event, emotion and moral. Classification gold labels are binned by score > 2.5.}
    \label{tab:llama3-zs-results}
    \vspace{-1em}
\end{table*}

\begin{figure}[ht!]
    \centering
    \resizebox{0.8\columnwidth}{!}{
        \includegraphics{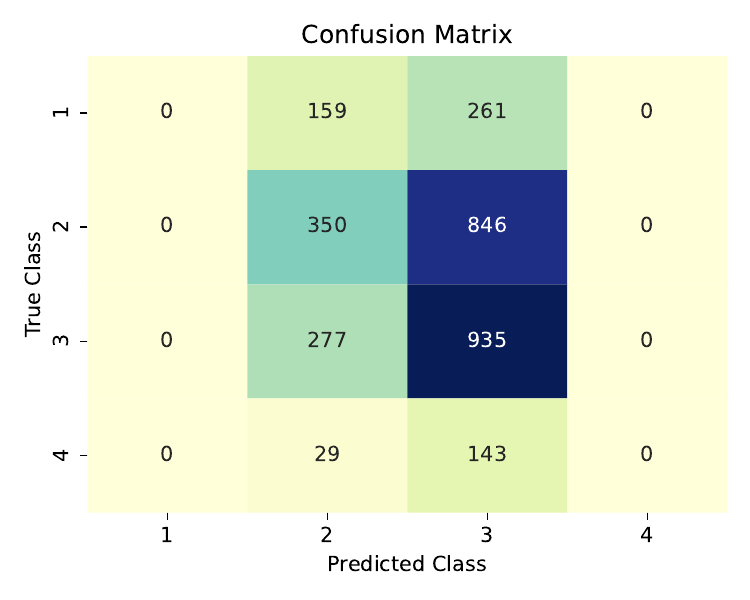}
    }
    \caption{Confusion matrix of fine-tuned Llama-3-8B on training set. The model could not estimate the corresponding score conditioned on the input story pair, but sampled a similarity class based on \textit{the gold class distribution of training data $P(Y)$ = (0.140, 0.399, 0.404, 0.057)} whatever the input pair was, leading to randomness on seen and unseen cases prediction. It learned nothing but statistical $P(Y)$ of training set. See more in \secref{sec:understandbottleneck}.}
    \label{fig:cm}
\end{figure}

\clearpage
\paragraph{Understanding Bottleneck of Fine-tuned LLMs.} \figref{fig:cm} shows the confusion matrix predicted by fine-tuned Llama-3-8B on training set. The model learned nothing but statistical $P(Y)$ of training set, leading to random predictions conditioned on input story pairs. See more in \secref{sec:understandbottleneck}.

\clearpage
\section{Annotation Guidelines}
\label{annotators_instruction}
This figure shows the guidelines provided to annotators for obtaining annotations on the Roman Urdu dataset. Annotators were first familiarized with established tasks in the English language before being introduced to the Urdu dataset, which is their native language.

\begin{figure}[ht!]
    \centering
        \includegraphics[scale=0.6]{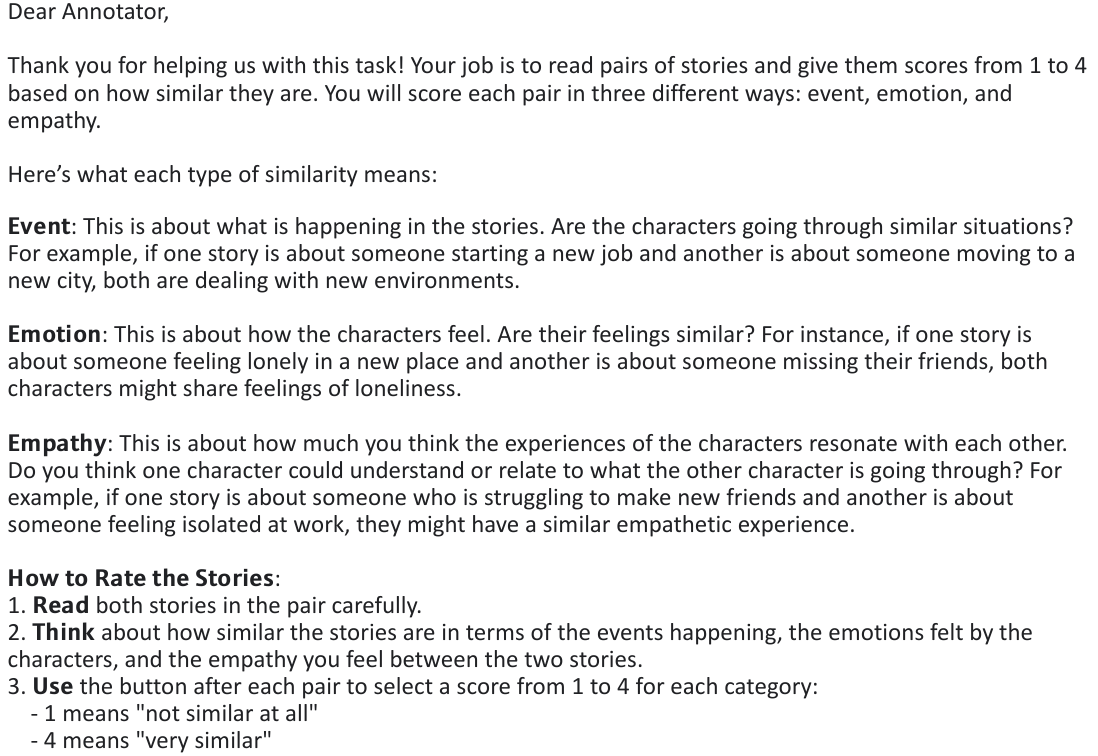}
    \caption{Annotation guidelines provided to annotators.}
    \label{fig:instruction}
\end{figure}

\paragraph{Recruitment And Payment}
Four Urdu native speakers were employed to annotate Urdu dataset. We pay 1 AED per story pair for each annotator, in total of 300 * 4 = 1200 AED.

\clearpage
\section{Urdu Story Pair Generation}
\begin{figure}[ht!]
    \centering
        \includegraphics[scale=0.5]{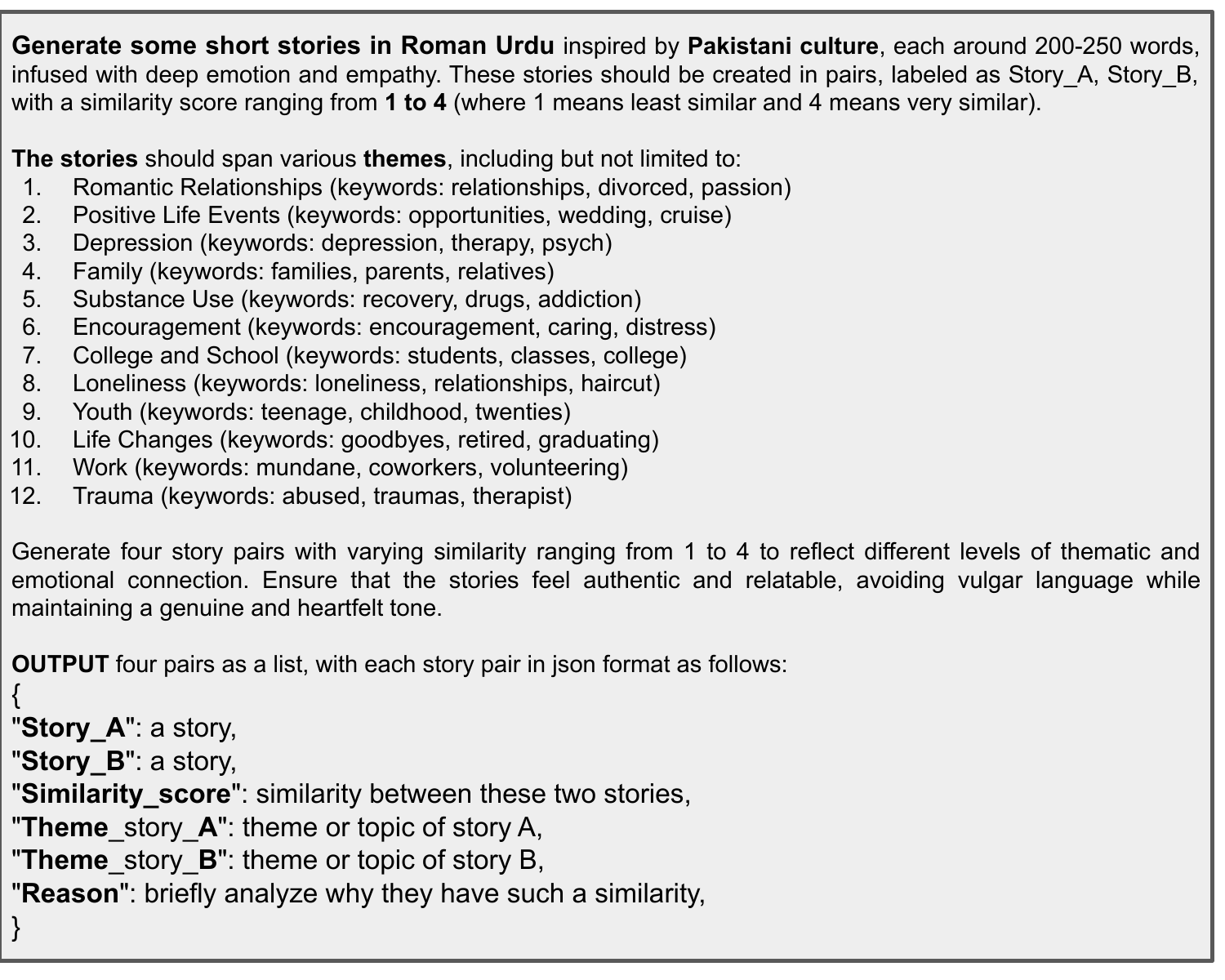}
    \caption{Prompt used for Urdu story pair generation based on GPT-4o.}
    \label{fig:urdustoryprompt}
\end{figure}

\begin{figure}[ht!]
    \centering
        \includegraphics[scale=0.7]{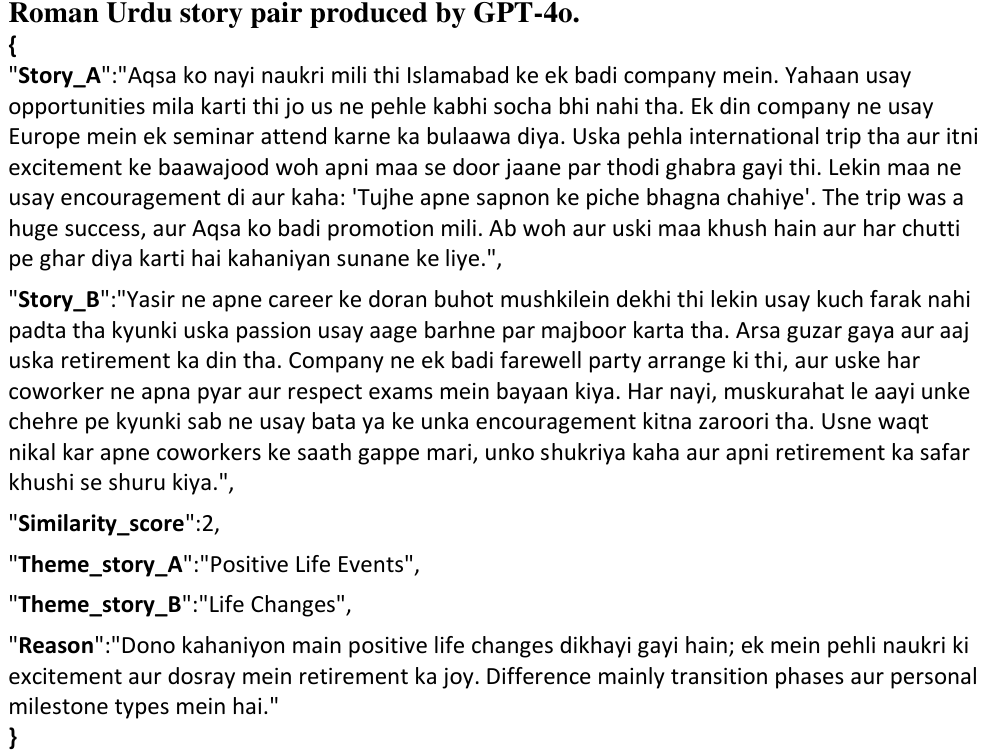}
    \caption{Example of Urdu story pair.}
    \label{fig:urdu_sample}
\end{figure}

\figref{fig:urdustoryprompt} displays the prompt we prepared to synthetically generate the Roman Urdu story data. To maintain consistency between the English and Urdu datasets, we incorporated all the themes mentioned in \cite{shen-etal-2023-modeling}. \figref{fig:urdu_sample} shows one of the 300 story pairs produced by \gptfouro. 

\clearpage
\section{English Story Pair Example}
\label{englis_pair_example}
The illustrative example form the EmpathicStories dataset \cite{shen-etal-2023-modeling} shows a sample pair of stories along with their summarized versions and scores in four aspects: empathy, event, emotion, and moral. 
\begin{figure}[ht!]
    \centering
        \includegraphics[scale=0.7]{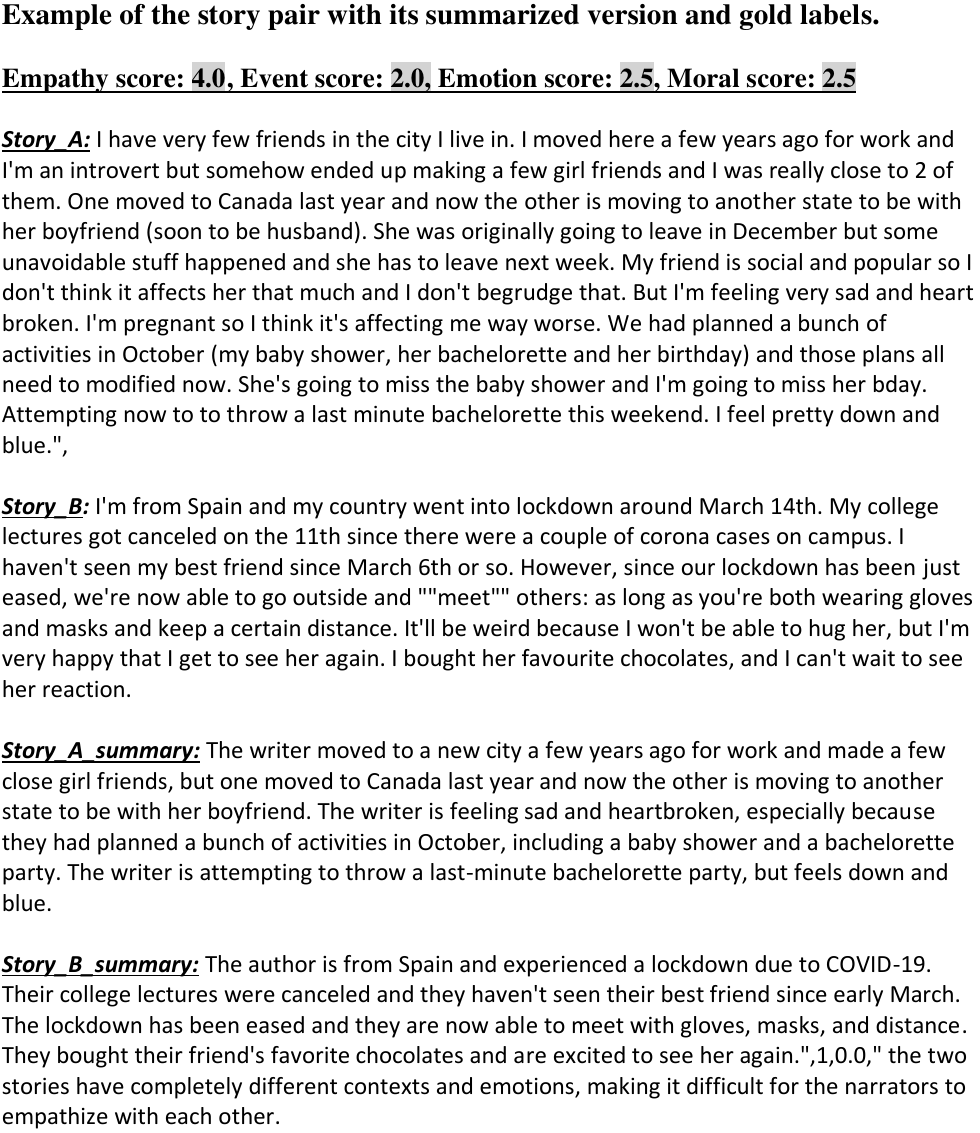}
    \caption{Example of English story pair.}
    \label{fig:english_sample}
\end{figure}

\end{document}